\documentclass[10pt,twocolumn,letterpaper]{article}

\usepackage[pagenumbers]{cvpr} 

\usepackage{graphicx}
\usepackage{amsmath}
\usepackage{amssymb}
\usepackage{booktabs}
\usepackage{enumitem}
\usepackage{multirow}
\usepackage{xcolor}
\usepackage{adjustbox}
\usepackage[toc,page]{appendix}
\usepackage{caption}
\usepackage{comment}
\usepackage{subcaption}
\usepackage{cite}
\usepackage[colorlinks]{hyperref}
\usepackage{pifont}
\usepackage{xspace}
\usepackage{algpseudocode}

\usepackage{booktabs}

\usepackage{blindtext}

\makeatletter
\DeclareRobustCommand\onedot{\futurelet\@let@token\@onedot}
\def\@onedot{\ifx\@let@token.\else.\null\fi\xspace}

\def\eg{\emph{e.g}\onedot} 
\def\ie{\emph{i.e}\onedot}

\def\etal{\emph{et al}\onedot}

\newcommand{\twowidth}{0.49}
 
\newcommand{\fourwidth}{0.24} 
\newcommand{\fivewidth}{0.19}

\newcommand{\cmark}{\ding{51}}%
\usepackage[capitalize]{cleveref}
\crefname{section}{Sec.}{Secs.}
\Crefname{section}{Section}{Sections}
\Crefname{table}{Table}{Tables}
\crefname{table}{Tab.}{Tabs.}

\def\cvprPaperID{3546} 
\def\confName{CVPR}
\def\confYear{2022}

\begin{document}

\title{Learning Spatio-Temporal Downsampling for Effective Video Upscaling}

\author{Xiaoyu Xiang$^1$, Yapeng Tian$^2$\thanks{This work is done during Yapeng's internship at Meta.}, Vijay Rengarajan$^1$, Lucas Young$^1$, Bo Zhu$^1$, Rakesh Ranjan$^1$ \\
$^{1}$Meta Reality Labs, $^{2}$University of Rochester \\
 \tt\small \{xiangxiaoyu,apvijay,lucasyoung,bozhufrl,rakeshr\}@fb.com, yapengtian@rochester.edu}

\maketitle

\begin{abstract}
Downsampling is one of the most basic image processing operations.
Improper spatio-temporal downsampling applied on videos can cause aliasing issues such as moir\'{e} patterns in space and the wagon-wheel effect in time. 
Consequently, the inverse task of upscaling a low-resolution, low frame-rate video in space and time becomes a challenging ill-posed problem due to information loss and aliasing artifacts. 
In this paper, we aim to solve the space-time aliasing problem by learning a spatio-temporal downsampler. 
Towards this goal, we propose a neural network framework that jointly learns spatio-temporal downsampling and upsampling. 
It enables the downsampler to retain the key patterns of the original video and maximizes the reconstruction performance of the upsampler. 
To make the downsamping results compatible with popular image and video storage formats, the downsampling results are encoded to uint8 with a differentiable quantization layer.
To fully utilize the space-time correspondences, we propose two novel modules for explicit temporal propagation and space-time feature rearrangement. 
Experimental results show that our proposed method significantly boosts the space-time reconstruction quality by preserving spatial textures and motion patterns in both downsampling and upscaling. 
Moreover, our framework enables a variety of applications, including arbitrary video resampling, blurry frame reconstruction, and efficient video storage. 
\end{abstract}

\section{Introduction}

\begin{figure}[t]
\captionsetup[subfigure]{labelformat=empty}
\begin{center}
  \begin{subfigure}[b]{\twowidth\linewidth}
  \includegraphics[width=\linewidth]{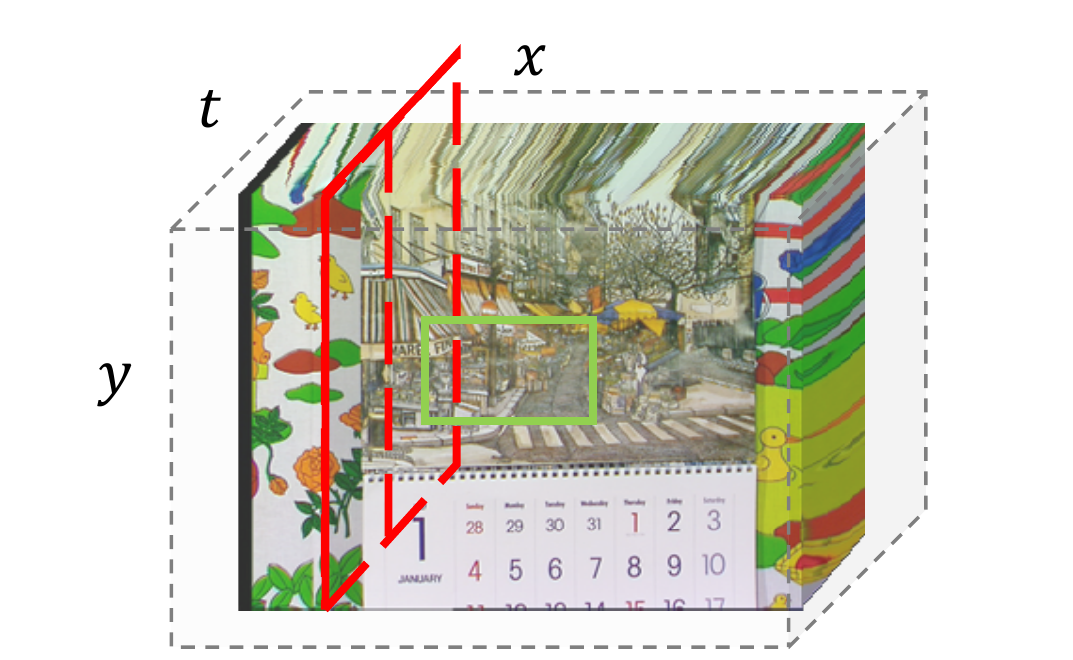}
  \subcaption{$xyt$ volume}
  \end{subfigure}
 \begin{subfigure}[b]{\twowidth\linewidth}
  \includegraphics[width=\linewidth]{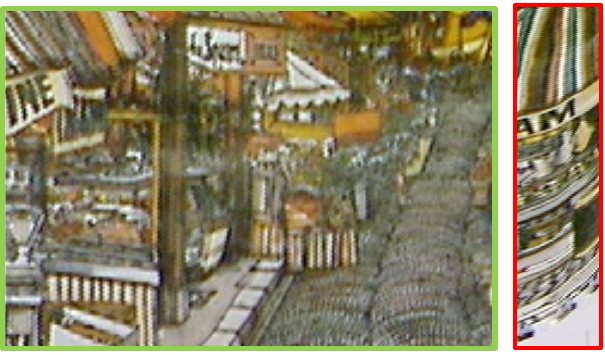}
  \subcaption{GT \textcolor{green}{image}/\textcolor{red}{temporal profile}}
  \end{subfigure}  
  
 \begin{subfigure}[b]{\twowidth\linewidth}
 \includegraphics[width=\linewidth]{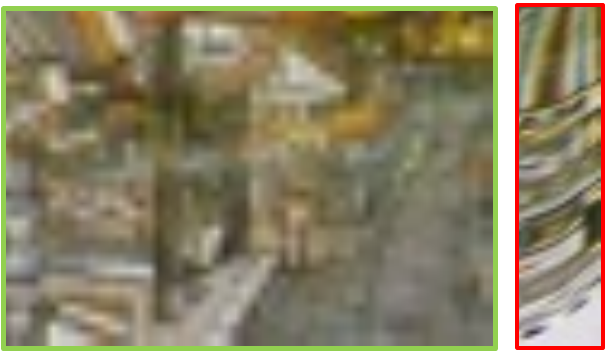}
  \subcaption{Bicubic+Nearest}
  \end{subfigure}
\begin{subfigure}[b]{\twowidth\linewidth}
 \includegraphics[width=\linewidth]{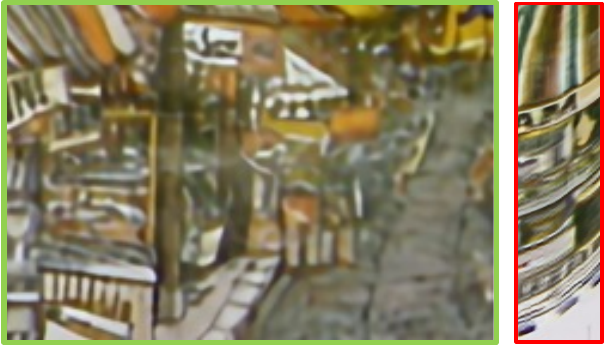}
  \subcaption{ZSM-Output~\cite{xiang2021zooming}}
  \end{subfigure}
  
\begin{subfigure}[b]{\twowidth\linewidth}
 \includegraphics[width=\linewidth]{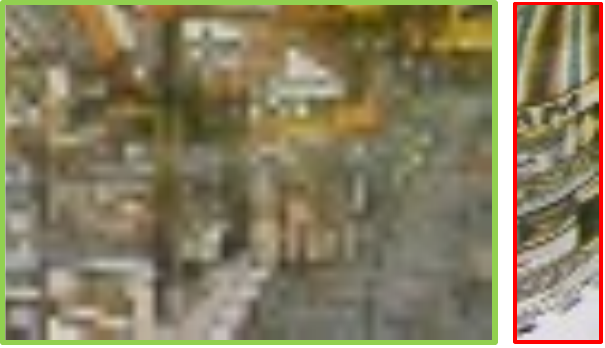}
  \subcaption{Ours-Downsample}
  \end{subfigure}
\begin{subfigure}[b]{\twowidth\linewidth}
 \includegraphics[width=\linewidth]{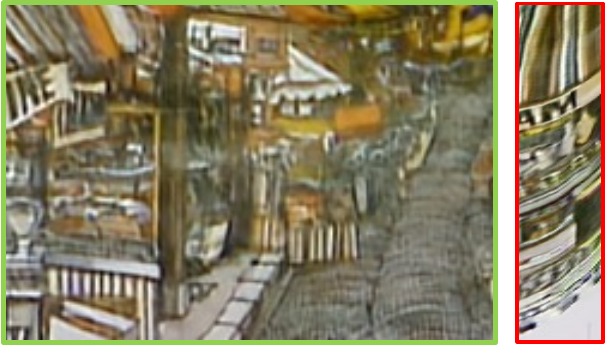}
  \subcaption{Ours-Upscale}
  \end{subfigure}
\end{center}
\caption{\textit{Effect of learned downsampling and upscaling for space-time reconstruction.} Compared to previous methods, the outputs of our learned downsampler maintain better spatial-temporal patterns, thus leading to more visually-appealing reconstruction results with better space-time consistency.}
\label{fig:teaser}
\end{figure}

Resizing is one of the most commonly used operations in digital image processing. Due to the limit of available memory and transfer bandwidth in compact devices, \eg mobile phones and glasses, the high resolutions, high frame rate videos captured by such devices trade off either spatial or temporal resolution~\cite{allebach1996edge,xiang2020boosting}. While nearest-neighbor downsampling is the standard operation to perform such reduction in space and time, it is not the best option: it folds over high-frequency information in the downsampled frequency domain, leading to aliasing as indicated by the Nyquist theorem~\cite{Shannon1949}. One way to avoid aliasing is to deliberately smudge high-frequency information by allowing more space and time to capture a single sample, \ie by optical blur~\cite{keelan2002handbook,ray1999scientific} and motion blur~\cite{korein1983temporal,shinya1993spatial,dachille2000high}, respectively. These spatial and temporal anti-aliasing filters band-limit frequencies, making it possible to reconstruct fine details during post-capture. Pre-designed anti-aliasing filters can be employed with the downsampler during capture time; for instance, using an optical low-pass filter~\cite{suzuki_1987} for spatial blur, and computational cameras such as the flutter shutter~\cite{raskar2006coded} camera for temporal blur. 

To design the optimal filter for a specific task, it is natural to incorporate the downstream performance in the loop, where the weights of the downsampling filter are updated by the objective function of the task~\cite{Talebi_2021_ICCV}. Kim~\etal~\cite{kim2018task} inverts the super-resolution network as a downscaling encoder. Zou~\etal~\cite{zou2020delving} proposes to adaptively predict filter weights for each spatial location. 

For the video restoration task, a major benefit of pre-designing the downsampler is to allow the co-design of an upsampler that recovers missing high-frequency details. While traditional methods mainly focus on upsampling -- super-resolution (SR) in the case of space, and video frame interpolation (VFI) in the case of time -- they assume the downsampler to be a trivial module. The LR images for SR tasks are usually acquired by bicubic downsampling.
For temporal reconstruction tasks like VFI, the low-fps (frames per second) frames are acquired by nearest-neighbor sampling, which keeps one frame per interval. Obviously, these operations are not optimal - the stride in time will lead to temporal aliasing artifacts.
This independent tackling of the upsampling stage makes solving the inverse problem harder, and it is typical to employ heavy priors on texture and motion, which results in inaccurate hallucination of lost details. On the other hand, jointly handling upsampling together with downsampling would enable better performance in retaining and recovering spatio-temporal details. In this paper, we explore the design of a joint framework through simultaneous learning of a downsampler and an upsampler that effectively captures and reconstructs high-frequency details in both space and time.

Based on the above observations, we propose a unified framework that jointly learns spatio-temporal downsampling and upsampling, which works like an auto-encoder for low-fps, low-resolution frames. 
To handle the ill-posed space-time video super-resolution problem, we first make the downsampler to find the optimal representation in the low-resolution, low-fps domain that maximizes the restoration performance. Moreover, considering the downsampled representation should be stored and transmitted in the common image and video data formats, we quantize them to be uint8 with a differentiable quantization layer that enables end-to-end training. Finally, the downsampled frames are upscaled by our upsampler. To improve the reconstruction capability, we devise space-time pixel-shuffle and deformable temporal propagation modules to better exploit the space-time correspondences. 

The main contributions of our paper are summarized as follows:

1. We provide a new perspective for space-time video downsampling by learning it jointly with upsampling, which 
preserves better space-time patterns and boosts restoration performance.

2. We observe that naive 3D convolution cannot achieve high reconstruction performance, and hence, we propose the deformable temporal propagation and space-time pixel-shuffle modules to realize a highly effective model design.

3. Our proposed framework exhibits great potential to inspire the community. We discuss the following applications: video resampling with arbitrary ratio, blurry frame reconstruction, and efficient video storage.



\section{Related Works}

\subsection{Video Downsampling}
\noindent
\textbf{Spatial.} Spatial downsampling is a long-standing research problem in image processing. 
Classical approaches, such as the box, nearest, bicubic, and Lasnczos~\cite{keys1981cubic,reinhard2010high,duchon1979lanczos}, usually design image filters to generate low resolution (LR) images by removing high frequencies and mitigating aliasing. Since most visual details exist in the high frequencies, they are also removed during downsampling. To address the problem, a series of structure and detail-preserving image downscaling methods~\cite{kopf2013content,oeztireli2015perceptually,weber2016rapid} are proposed. Although these approaches can produce visually appealing LR images, they cannot guarantee that upscaling methods can restore the original high resolution (HR) images due to aliasing and non-uniform structure deformation in the downscaled LR images. Pioneering research~\cite{triggs2001empirical,trentacoste2011blur} demonstrates that it is possible to design filters that allow for reconstructing the high-resolution input image with minimum error. The key is to add a small amount of optical blurring before sampling. Inspired by this, we propose to automatically learn the best blurring filters during downsampling for more effective visual upscaling.

\noindent
\textbf{Temporal.} A simple way to downsample along the temporal dimension is to increase the exposure time of a frame and capture the scene motion via blur. The loss of texture due to averaging is traded off for the ability to embed motion information in a single frame. Blur-to-video methods~\cite{jin2018learning, purohit2019bringing, jin2019learning, shen2020blurry, zhang2020every, argaw2021motion} leverage motion blur and recover the image sequence, constraining the optimization with spatial sharpness and temporal smoothness. To regularize the loss of texture during capture, Yuan~\etal~\cite{yuan2007image} use a long and short exposure pair for deblurring, while Rengarajan~\etal~\cite{rengarajan2020photosequencing} exploit the idea to reconstruct high-speed videos. Coded exposure methods replace the box-filter averaging over time with a broadband filter averaging by switching the shutter on and off multiple times with varying on-off durations within a single exposure period. This results in better reconstruction owing to the preservation of high-frequency details over space. Raskar~\etal~\cite{raskar2006coded} use such a coded exposure camera for deblurring, while Holloway~\etal~\cite{holloway2012flutter} recover a high-speed video from a coded low frame rate video. Our work contributes an extension of this previous work by learning the optimal temporal filters during downsampling for restoring sharp high-fps videos.

\subsection{Video Upscaling}
\noindent
\textbf{Video Super-Resolution.} The goal of video super-resolution (VSR) is to restore HR video frames from their LR counterparts. Due to the existence of visual motion, the core problem to solve in VSR is how to temporally align neighboring LR frames with the reference LR frame.  
Optical flow methods seek to compute local pixel shifts and capture motions. Thus, a range of VSR approaches~\cite{caballero2017real,tao2017detail,sajjadi2018frame,wang2018learning,xue2019video,haris2019recurrent} use optical flow to estimate motion and then perform motion compensation with warping. However, optical flow is generally limited in handling large motions, and flow warping can introduce artifacts into aligned frames. To avoid computing optical flow, implicit temporal alignment approaches, such dynamic upsampling filters~\cite{jo2018deep}, recurrent propagation~\cite{lim2017deep,huang2017video,isobe2020video}, and deformable alignment~\cite{tian2018tdan,wang2019edvr,chan2020understanding,chan2021basicvsr++} are utilized to handle complex motions.

\noindent
\textbf{Video Frame Interpolation.} Video frame interpolation (VFI) aims to synthesize intermediate video frames in between the original frames and upscale the temporal resolution of videos. Meyer~\etal~\cite{meyer2015phase} utilizes phase information to assist frame interpolation. 
\cite{long2016learning,kalluri2020flavr} proposed an encoder-decoder framework to directly predict intermediate video frames. Niklaus~\etal~\cite{niklaus2017adavonv,niklaus2017adsconv} utilizes a spatially-adaptive convolution kernel for each pixel for synthesizing missing frames.
Similar to VSR, optical flow is also adopted in VFI approaches~\cite{jiang2018super,liu2017video,niklaus2018context, bao2019memc, bao2019depth,sim2021xvfi} to explicitly handle motions.

\noindent
\textbf{Space-Time Video Super-resolution} The pioneering work to extend SR to both space and time domains was proposed by Shechtman \etal~\cite{shechtman2002increasing}. Compared with VSR and VFI tasks, STVSR is even more ill-posed since pixels are missing along both spatial and temporal axes. To constrain the problem, early approaches~\cite{shechtman2002increasing,mudenagudi2010space,takeda2010spatiotemporal,shahar2011space} usually exploited space-time local smoothness priors. Very recently, deep learning-based STVSR frameworks~\cite{kim2020fisr,xiang2020zooming,haris2020space,xiao2020space,dutta2021efficient,xiang2021zooming,xu2021temporal} have been developed. These methods directly interpolate the missing frame features and then upsample all frames in space. Like the video frame-interpolation methods, the input frames are anchors of timestamps, which limits the upscale ratio in the temporal dimension. Unlike these methods, we aim to freely resize the space-time volume with arbitrary scale ratios in this work.  

\section{Space-Time Anti-Aliasing (STAA)}

We first explain the intuition behind STAA: treating video as spatio-temporal $xyt$ volume and leveraging the characteristic spatio-temporal patterns for video reconstruction. Towards this end, we propose efficiently utilizing the spatio-temporal patterns in upscaling with space-time pixel-shuffle. However, naively regarding time as an additional dimension beyond space has a limitation: the pixel $(x,y)$ at the $i$-th frame is usually not related to the same $(x,y)$ at the $i+k$-th frame. To tackle this problem, we model the temporal correspondences with deformable sampling.

\subsection{Intuition}
To tackle the aliasing problem, previous methods insert low-pass filters either in space or time. However, we noticed that the space and time dimensions of a $xyt$ volume are not independent: as shown in Fig.~\ref{fig:teaser}, the temporal profiles $xt$ and $yt$ display similar patterns as the spatial patches. In an ideal case where the 2D objects move with a constant velocity, the temporal profile will appear as a downsampled version of the object~\cite{shahar2011space,pollak2020across}, as illustrated in Fig.~\ref{fig:fourier}. This space-time patch recurrence makes it possible to aid the reconstruction of the under-sampled dimension with abundant information from other dimensions. Based upon this observation, we adopt a 3D low-pass filter on both space and time to better utilize the correspondences across dimensions. Accordingly, our upsampler network also adopts the 3D convolutional layers as the basic building block due to its capability of jointly handling the spatio-temporal features.

\subsection{Module Design}
\label{sec:module_design}
\begin{figure}[tb]
     \centering
     \includegraphics[width=\linewidth]{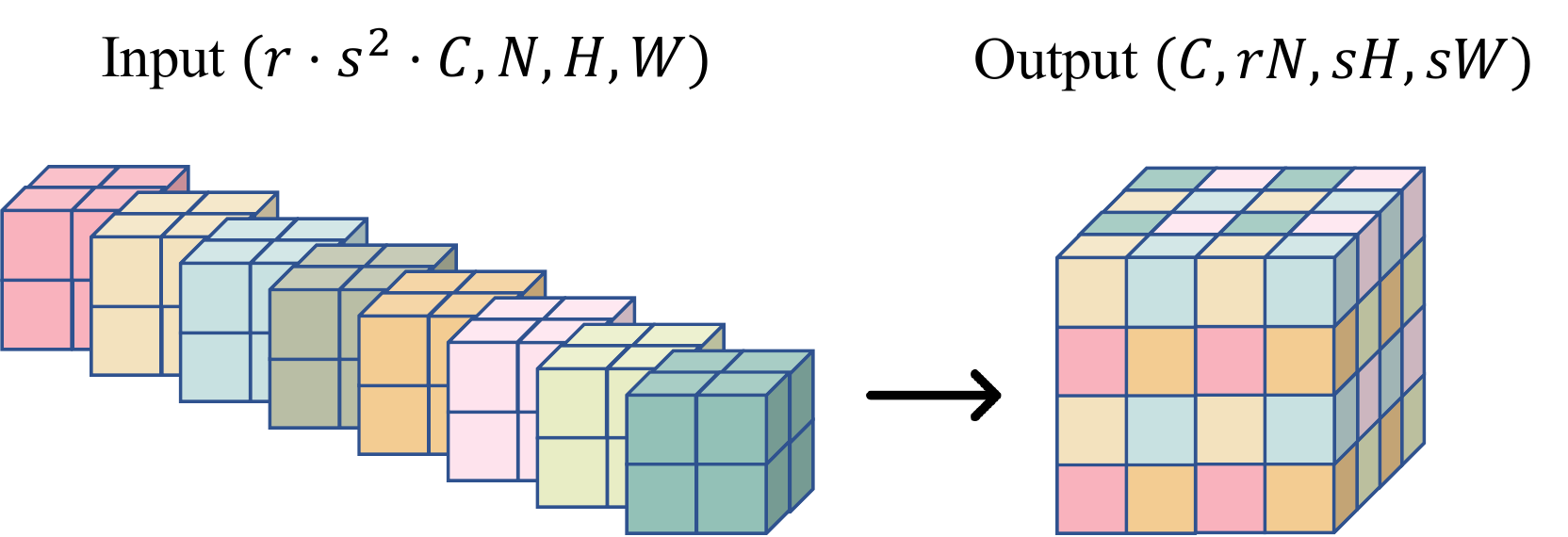}
\caption{Space-Time Pixel-Shuffle.}
\label{fig:st-shuffle}
\end{figure}

\textbf{Space-Time Pixel-Shuffle.} Pixel-shuffle~\cite{shi2016real} is a widely-used operation in single image super-resolution (SISR) for efficient subpixel convolution. It has two advantages: the learned upscaling filter can achieve the optimal performance; computational complexity is reduced by rearranging the elements in LR feature maps. Inspired by its success in SISR, we extend it to space-time. Fig.~\ref{fig:st-shuffle} illustrates the shuffling operation: for a input tensor with shape $(r\cdot s^2\cdot C, N, H, W)$, the elements are shuffled periodically into the shape $(C, rN, sH, sW)$. 

\noindent
\textbf{Naive Deconvolution Is Insufficient.} If we simply regard the space-time upscaling as the reverse process of the downsampling, then conceptually, a deconvolution should be enough to handle this process. To investigate this idea, we build a small network using 3D convolutions and the aforementioned space-time pixel-shuffle layers with a style of ESPCN~\cite{shi2016real}. Although this network does converge, it only improves the PSNR by $\sim 0.5$ dB compared with trilinear upscaling the $xyt$ cube -- such improvement is too trivial to be considered effective, particularly when compared to the success of ESPCN in SISR. 

\begin{figure}[tb]
     \centering
     \includegraphics[width=.95\linewidth]{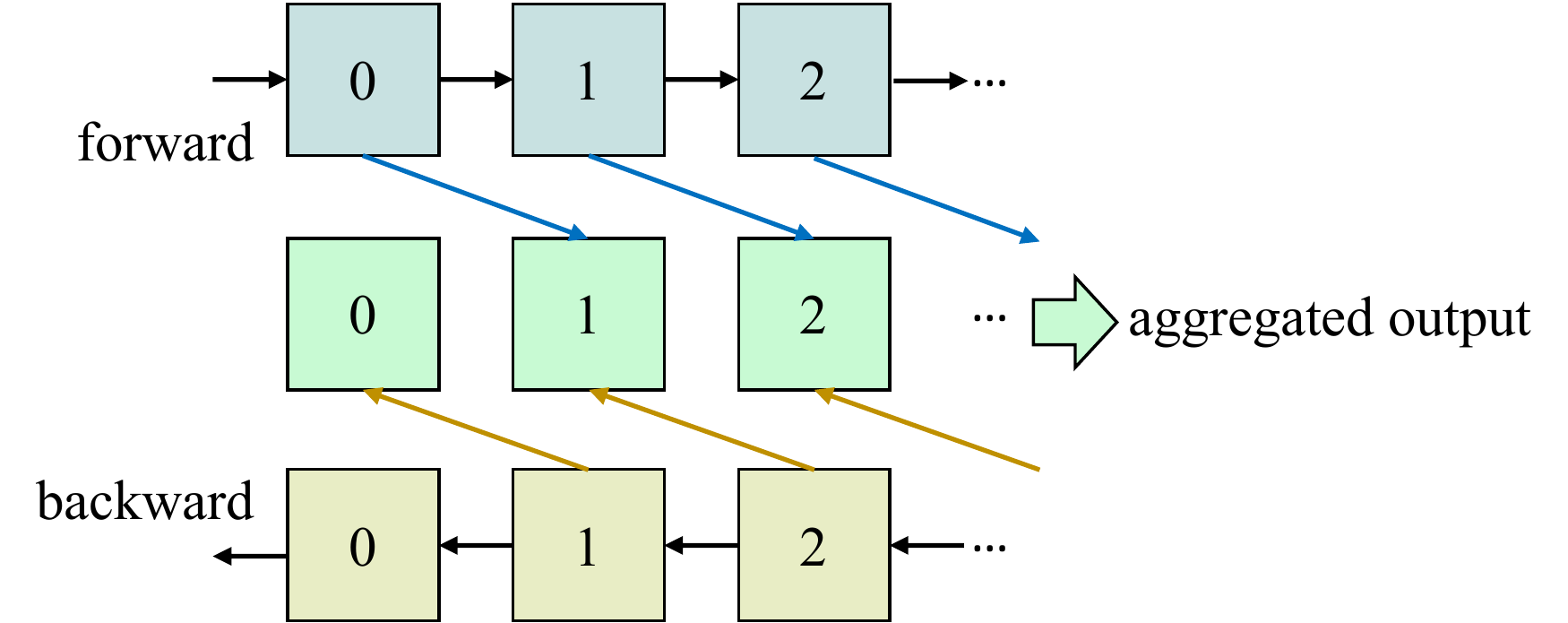}
\caption{Temporal propagation module.}
\label{fig:temp-model}
\end{figure}

\noindent
\textbf{Enhance Temporal Modeling Capacity.} As noted above, this suboptimal result was expected due to the lost correspondences between the $i$-th and $i+k$-th frames. Thus, 3D convolution alone cannot guarantee a good reconstruction performance due to its relatively small field of view.
Understanding ``what went where"~\cite{josh2003what} is the fundamental problem for video tasks. Such correspondence is even more critical in our framework: our STAA downsampler encodes the motion by dispersing the space feature along the temporal dimension. Correspondingly, during the reconstruction stage, the supporting information can come from neighboring frames. Motivated by this, we devise a deformable module to build temporal correspondences and enhance the model's capability to handle dynamic scenes: for a frame at time $i$, it should look at adjoining $i-k,\ldots i+k$ frames and refine the current feature by aggregating the relevance. For efficient implementation, we split the information propagation into forward and backward directions, where the temporal correspondence is built and passed recurrently per direction, as shown in Fig.~\ref{fig:temp-model}. The refined features from both forward and backward passes are aggregated to yield the output. Hence, the difficulty of perceiving long-range information within the 3D convolutional receptive field is alleviated.

\section{Joint Downscaling and Upscaling Framework}
\begin{figure*}[tbp]
\centering
\includegraphics[width=\linewidth]{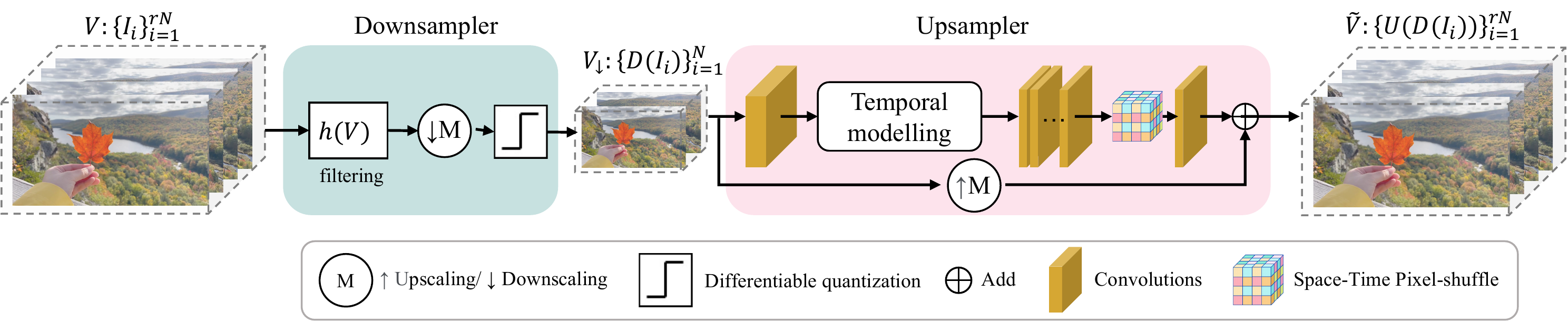}
\caption{Our training framework functions as an auto-encoder in which we train the downsampler $D$ (encoder), and the upsampler $U$ (decoder) jointly in an end-to-end manner.} 
\label{fig:framework}
\end{figure*}

Our framework architecture is shown in Fig.~\ref{fig:framework}: given a sequence of video frames $V=\{I_i\}_1^{rN}$ where each $I_i$ is an RGB image of dimensions $sW \times sH$ ($r$ and $s$ as scale factors), our goal is to design (a) a downsampler which would produce $V_\downarrow=\{D(I_i)\}_1^{N}$ where each $D(I_i)$ has the dimensions $W \times H$, and, (b) an upsampler which would produce $\widetilde{V}=\{U(D(I_i))\}_1^{rN}$ where, in the perfect case, $V = \widetilde{V}$.

\subsection{Downsampler}
\label{subsec: downsampler}

Our downsampler consists of a 3D low pass filter $h(\cdot)$ followed by a downsampling operation by striding.
Given a sequence of input frames $V=\{I_i\}_1^{rN}$ that needs to be downsampled, 
we first convolve it with the filter $h$ as follows:
\begin{equation}
    h(V)[t,x,y] = \sum_{i,j,k\in\Omega}h[i,j,k]\cdot V[t-i, x-j, y-k].
\end{equation}

To ensure that the learned filters are low-pass, we add a softmax layer to regularize the weight values within the range $[0, 1]$ and the sum to be 1. We then use striding to produce our desired downsampled frames in both space and time. An ideal anti-aliasing filter should restrict the bandwidth to satisfy the Nyquist theorem without distorting the in-band frequencies.

\begin{figure*}[tbp]
\centering
\includegraphics[width=\linewidth]{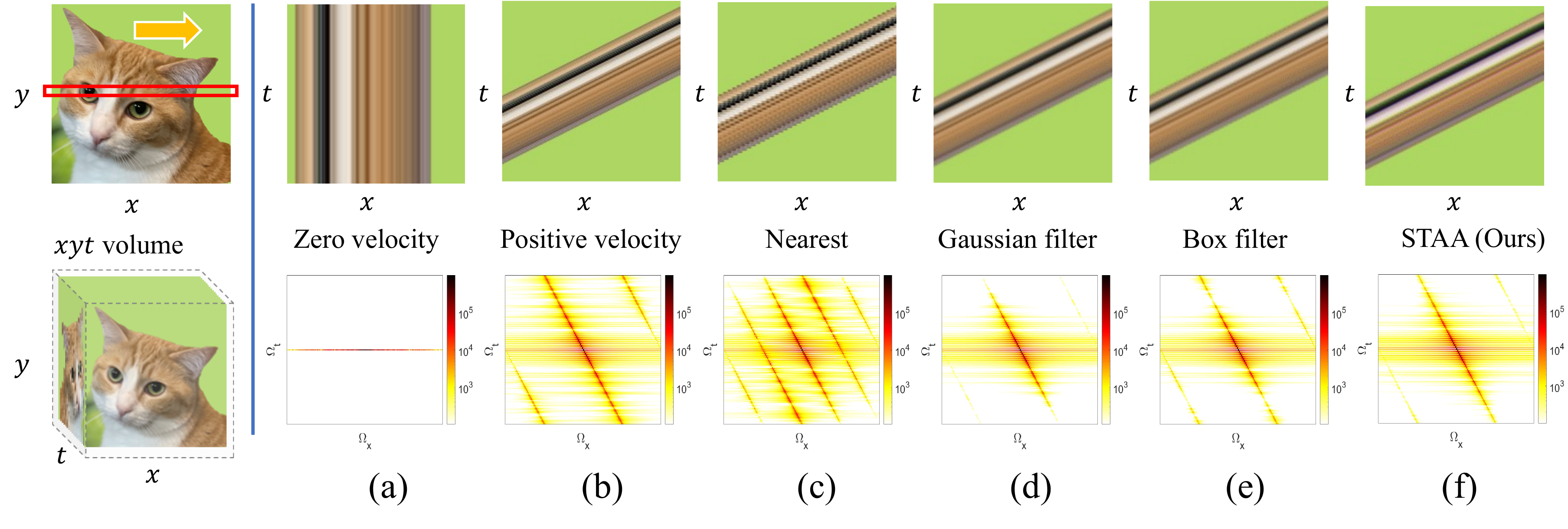}
\vspace{-4mm}
\caption{Space-time and Fourier domain plots for a moving object.} 
\label{fig:fourier}
\vspace{-3mm}
\end{figure*}

\noindent
\textbf{Analysis of learned filters.}
We present a study based on the frequency domain analysis of spatio-temporal images to compare different types of low-pass filters. 
We analyze the canonical case of a single object moving with uniform velocity. 
The basic setup is shown in the first column of Fig.~\ref{fig:fourier}, where the top row shows the static object and the bottom row shows the $xyt$ volume corresponding to the motion. 
Figs.~\ref{fig:fourier}(a) to (f) show the temporal profiles $xt$ corresponding to the 1D scanline (marked in red) for various scenarios/filters in the top row and their corresponding Fourier domain plots in the bottom row.
Please check the supplementary material for more details about how the Fourier plots are calculated and what the spectra components represent.

In Fig.~\ref{fig:fourier}, (a) is a space-time diagram for a static scene (zero velocity), so there is no change along the time (vertical) dimension. In (b), we can see that the motion causes a time-varying effect, which results in shear along the spatial $x$ direction. This shows the coupling of spatial and temporal dimensions. 
Applying just the nearest-neighbor downsampling in time leads to severe aliasing, as shown by duplication of streaks in (c) bottom row. Thus, the plain downsampling method leads to temporal pattern distortions. Characteristic spatio-temporal patterns relate to events in the video~\cite{niyogi1994analyzing,zelnik2001event,cooper2007video}. Thus, good downsampling methods should also retain the ``textures" in time dimension. 

Figs.~\ref{fig:fourier}(d), (e), and (f) show the images and frequency plots for the case of applying Gaussian, box, and our STAA low pass filters, respectively, first, followed by nearest-neighbor downsampling. 
%
The convolution with these filters causes blurring across space and time dimensions, as shown in the images, and since convolution corresponds to frequency domain multiplication, we can see the benefits of these filters visually in the frequency domain plots. 
The Gaussian filter reduces the subbands and high spatiotemporal frequencies in (d). 
The Box filter along the temporal dimension, which is used to describe the motion blur~\cite{egan2009frequency,brooks2019learning,rim2022realistic}, destroys spatial details and attenuates certain temporal frequencies, which causes post-aliasing~\cite{mitchell1988reconstruction} during reconstruction. 
Our proposed STAA filter attenuates high frequencies and the subbands like the Gaussian filter does, but at the same time, it preserves more energy in the main spectra component, as shown in (f). This characteristic ensures the prefiltered image maintains a good spatio-temporal texture, thus benefiting the reconstruction process. 

\noindent
\textbf{Connection to coded exposure.} 
%
Traditional motion blur caused by long exposure can be regarded as filtering with a temporal box filter. While long exposure acts as a natural form of filtering during the capture time itself, the box filter is not the best filter for alias-free reconstruction~\cite{mitchell1988reconstruction}. Hence, coded exposure methods ``flutter" the shutter in a designed sequence to record the motion without completely smearing the object across the spatial dimension to recover sharp high-frame-rate images and videos~\cite{raskar2006coded, holloway2012flutter}. Our learned downsampler can be regarded as a learned form of the coded exposure: considering the temporal kernel size as an exposure window, we aggregate the pixels at each time step to preserve an optimal space-time pattern for better reconstruction.


\noindent
\textbf{Differentiable quantization layer.}
The direct output of our downsampler is a floating-point tensor, while in practical applications, images are usually encoded as 8-bit RGB (uint8) format. Quantization is needed to make our downsampled frames compatible with popular image storage and transmission pipelines. However, this operation is not differentiable. The gap between float and discrete integer causes training unstable and a drop in performance. To bridge the gap, we adopt a differentiable quantization layer that enables end-to-end training. More details can be found in \textit{supplementary material}. 

\subsection{Upsampler}

Given a sequence of downsampled frames, $V_{\downarrow} = \{D(I_i)\}^{N}_{i=1}$, the upsampler $U$ aims to increase the resolution in both space and time. 
The estimated upscaled video $\widetilde{V}=\{U(D(I_i)\}_{i=1}^{rN}$ should be as close to the original input as possible.

To achieve this purpose, we choose 3D convolution as our basic building block for the upsampler. The input sequence is converted to the feature domain $\mathcal{F}$ by a 3D convolution. We adopt an deformable temporal modeling (DTM) subnetwork to aggregate the long-range dependencies recurrently. It takes the last aggregated frame feature $DTM(f_{i-1})$ at time step $i-1$ and the current feature $f_i$ as inputs, outputting the current aggregated feature:
\begin{equation}
    DTM(f_i) = T(f_i, DTM(f_{i-1})),
\end{equation}
where $f_i$ is the frame feature at time step $i$, and $T$ denotes a general function that finds and aligns the corresponding information to the current feature. We adopt the deformable sampling function~\cite{dai2017deformable,zhu2019deformable} as $T$ to capture such correspondences. To fully exploit the temporal information, we implement the DTM in a bidirectional manner that aggregates the refined features from both forward and backward passes.

The refined sequence is then passed to the reconstruction module that is composed of 3D convolutions. To fully explore the hierarchical features from these convolutional layers, we organize them into residual dense blocks~\cite{zhang2018residual}. It densely connects the 3D convolution layers into local groups and fuses the features of different layers. Following the previous super-resolution networks~\cite{lim2017enhanced}, no BatchNorm layer is used in our reconstruction module. 
Finally, a space-time pixel-shuffle layer is adopted to rearrange the features with a periodic shuffling across the $xyt$ volume~\cite{rogozhnikov2021einops}. 

We denote the output just after the space-time pixel-shuffle as $F(V_{\downarrow})$, where $F(\cdot)$ is all the previous operations for upscaling the input $V_{\downarrow}$. To help the main network focus on generating high-frequency information, we bilinearly upscale the input sequence and add it to the reconstructed features as the final output:
\begin{equation}
    U(V_{\downarrow}) = V_{\downarrow}\uparrow_M + F(V_{\downarrow}).
\end{equation}
This long-range skip-connection allows the low-frequency information of the input to bypass the major network and makes the major part of the network predict the residue. It ``lower-bounds" the reconstruction performance and increases the convergence speed of the network.

\section{Experiments}
First, we compare the proposed STAA framework to the state-of-the-art methods for upscaling videos in space (VSR) and time(VFI) to show the benefit of the jointly learned downsampling and upscaling. We also conduct ablation studies on the downsampler and upsampler designs. Finally, we discuss several potential applications of our method.

\noindent
\textbf{Datasets.} We adopt the Vimeo-90k dataset \cite{xue2019video} to train the proposed framework. It consists of more than 60,000 training video sequences, and each video sequence has seven frames. These frames are used as both the inputs and outputs of the auto-encoder. Consistent with previous works, we evaluate our method on the Vid4\cite{liu2011bayesian} dataset and the test set of Vimeo-90k to compare with the state-of-the-art VFI, VSR, and STVSR methods.

\noindent
\textbf{Implementation Details.}  In our proposed STAA framework, we adopt a filter of shape $3\times 3 \times 3$ for the downsampler. In the upsampler, one 3D convolutional layer is used to convert the input frame sequence to the feature domain. We adopt the deformable ConvLSTM~\cite{xiang2020zooming,xiang2021zooming} for temporal propagation. Five 3D residual-dense blocks are stacked to build the reconstruction module.
During the training phase, we augment the training frames by randomly flipping horizontally and $90^\circ$ rotations. The training patch size is $128\times128$, and the batch size is set to be 32. We train the network for 100 epochs using the Adam~\cite{kingma2014adam} optimizer. The initial learning rate is set to $2 \times 10^{-4}$ which scales down by a factor of $0.2$ each at epochs, 50 and 80. 
Our network is implemented in PyTorch~\cite{pytorch_neurips}. Our STAA framework is trained end-to-end with the $L$1-loss: $   L(V, \widetilde{V}) = ||V - \widetilde{V}||_1$.

\noindent
\textbf{Evaluation.} We use the Peak Signal-to-Noise-Ratio (PSNR) and Structural Similarity Index (SSIM)~\cite{wang2004image} metrics to evaluate video restoration. We also compare the number of parameters (million, M) to evaluate model efficiency.

\subsection{Comparison with State-of-the-Art Methods}

We compare the performance of reconstructing a video in space and time with the state-of-the-art methods in a cascaded manner: XVFI~\cite{sim2021xvfi} and FLAVR~\cite{kalluri2020flavr} for video frame-interpolation, and BasicVSR++~\cite{chan2021basicvsr++} for video super-resolution. We also compare with the one-stage ZSM~\cite{xiang2020zooming,xiang2021zooming} for space-time video super-resolution. Note that the previous methods cannot upsample by $2\times$ in time: for two input frames, they generate one interpolated frame along with the two inputs, while our STAA generates four upsampled ones. For an apples-to-apples comparison, we only calculate the PSNR/SSIM of the synthesized frames. Quantitative results on Vimeo-90k~\cite{xue2019video} and Vid4~\cite{liu2011bayesian} are shown in Table~\ref{tab:sota_comp}. 

Our method outperforms the previous methods by a large margin on all datasets and settings. For temporal upscaling, adopting the STAA downsampling and upscaling exceeds the second-best method by 8.28 dB on Vimeo-90k and 9.95 dB on Vid4, which validates the importance of anti-aliasing in the temporal dimension. For the challenging case of $4\times$ space/$2\times$ time, our method still demonstrates remarkable improvement by 2.4~dB on the Vid4 and more than 1~dB on the Vimeo-90k datasets. Such significant improvement brought by the co-design of downsampling filter and upscaling network provides a new possibility for improving current video restoration methods.

\begin{table*}[tbp]
\begin{center}
\caption{Comparisons with SOTA cascaded video frame interpolation (VFI) and super-resolution (VSR), and space-time super-resolution (STVSR) methods.}
\label{tab:sota_comp}
\resizebox{.9\linewidth}{!}{
\begin{tabular}{c|c|cc|c|cc|cc}
\hline
Upscale rate      & Downsampler                         & \multicolumn{2}{|c|}{Reconstruction Method}   & \multirow{2}{*}{Params/M}  & \multicolumn{2}{|c|}{Vimeo-90k}                        & \multicolumn{2}{|c}{Vid4}                            \\
time/space             & time/space                          & VFI   & VSR                                       &                           & PSNR & SSIM & PSNR & SSIM \\
\hline
\multirow{4}{*}{2$\times$/1$\times$} & \multirow{2}{*}{Nearest/-}    & XVFI~\cite{sim2021xvfi}  & - &             5.7                                  &           34.76                                     &        0.9532                  &       29.21                   &       0.9496                                           \\
                       &                                     & FLAVR~\cite{kalluri2020flavr} & - &        42.1 &     36.73                                  &    0.9632                                            &      29.83                    &    0.9585                                                                    \\ \cline{2-9}
                       & STAA                                 & \multicolumn{2}{c|}{Ours}     &          15.9                                     &       \textbf{45.01}                                         &     \textbf{0.9912}                     &                \textbf{39.78}          &          \textbf{0.9926}                                       \\ \hline
\multirow{6}{*}{2$\times$/4$\times$} & \multirow{3}{*}{Nearest/Bicubic} & XVFI~\cite{sim2021xvfi}  & BasicVSR++~\cite{chan2021basicvsr++}           &        5.7+7.3      &                32.41                 &         0.9123                                       &               24.90           &            0.7726                                                            \\
                       &                                     & FLAVR~\cite{kalluri2020flavr} & BasicVSR++~\cite{chan2021basicvsr++}           &     42.1+7.3                                          &         32.74                                       &            0.9119              &      24.79                    &        0.7678                                          \\
                       &                                     & \multicolumn{2}{c|}{ZSM~\cite{xiang2020zooming}}      &        11.1                                       &              33.48                                  &     0.9178                     &    24.82   
                      &         0.7763 \\
                     & & \multicolumn{2}{c|}{STDAN~\cite{wang2022stdan}}      &         8.3                                  &                           33.59                    &       0.9192                  &      24.91                     &     0.7832                                              \\ \cline{2-9}
                       & STAA                                 & \multicolumn{2}{c|}{Ours}     &        16.0    &                      \textbf{34.53}            &             \textbf{0.9426}                                   &      \textbf{27.31}                    &      \textbf{0.9173}                                        \\
                       \hline
\end{tabular}}
\end{center}
\vspace{-4mm}
\end{table*}

\subsection{Ablation Studies}
\textbf{Type of downsampling filter.}
To verify the effectiveness of our learned downsampling filter, we compare the reconstruction performance by switching downsampler types. The idea is: since the reconstruction capability of the same upsampler architecture is unchanged, a better reconstruction result means that the downsampler produces a better space-time representation. For an apples-to-apples comparison, we use the same kernel size for Gaussian and our STAA filters. We experiment different combinations of time/space sampling rates: $1\times t,2\times s$, $2\times t, 1\times s$, and $2\times t, 2\times s$.

As shown in Table~\ref{tab:abl_filter}, the nearest-bicubic downsampling, which is adopted by previous video reconstruction tasks, provides the worst representation among all. For the $2\times t/2\times s$ setting, the reconstruction network cannot converge to global optimal. Although it is still the dominant setting, it cannot handle the temporal-aliasing issue and might hinder the development of video reconstruction methods. Pre-filtering with a 3D Gaussian blur kernel can alleviate the aliasing problem, which exceeds the nearest-neighbor downsampling in time and the bicubic downsampling in space. Still, the Gaussian filter cannot produce the optimal spatio-temporal textures. Compared with these classical methods, our STAA filters improve the reconstruction performance by a large margin, as shown in the last four rows of each setting. We believe that our proposed STAA downsampler has the potential to serve as a new benchmark for video reconstruction method design and inspire the community from multiple perspectives.

\begin{figure}
\centering
\includegraphics[width=0.9\linewidth]{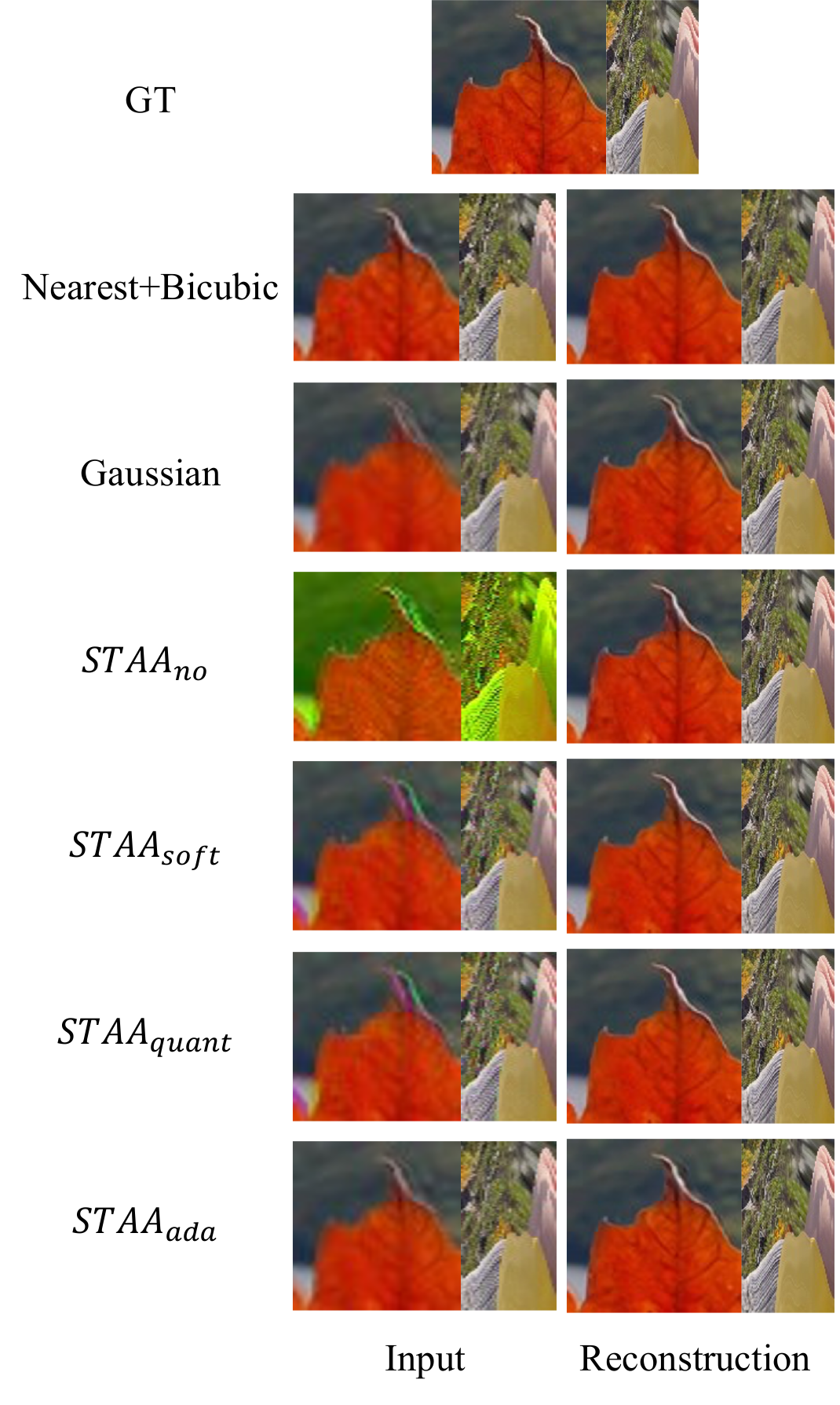}
\caption{Visual comparison at $2\times t/ 2\times s$ setting.}
\label{fig:comp_filter}
\vspace{-3mm}
\end{figure}

We visualize the downsampled frames and their corresponding reconstruction results in Fig.~\ref{fig:comp_filter}. The nearest-bicubic downsampled result looks nice per frame; however, the temporal profile has severe aliasing. In comparison, the anti-aliasing filters make the downsampled frames ``blurry" to embed the motion information. 

\begin{table}[t]
		\caption{Quantitative comparison of downsampling filters. We compare the reconstruction results of the classical nearest/bicubic sampling and Gaussian filter to our learned STAA filters with different constraints. The best two results are highlighted in \textcolor{red}{red} and \textcolor{blue}{blue}, respectively. }
    \label{tab:abl_filter}
		\centering
\resizebox{\linewidth}{!}{
\begin{tabular}{ccccc}
\hline
Downsampling rate & Time       & Space   & PSNR & SSIM \\ \hline
1$\times t$/2$\times s$             & -          & Bicubic & 41.61     &   0.9816   \\
                  & \multicolumn{2}{c}{Gaussian}            &   41.69   &   0.9827   \\
                  & \multicolumn{2}{c}{STAA$_{no}$}         &   43.47   &  0.9883    \\
                  & \multicolumn{2}{c}{STAA$_{soft}$}         &   \textcolor{red}{44.19}   &   \textcolor{red}{0.9901}   \\
                  & \multicolumn{2}{c}{STAA$_{quant}$}         &   \textcolor{red}{44.19}   &  \textcolor{blue}{0.9899}    \\
                  & \multicolumn{2}{c}{STAA$_{ada}$}         &   42.05   &   0.9840  \\
                  \hline
2$\times t$/1$\times s$             & Nearest &    -     &  38.22    &  0.9715    \\
                  & \multicolumn{2}{c}{Gaussian}            &   39.26   &  0.9796    \\
                  & \multicolumn{2}{c}{STAA$_{no}$}         &    46.50  &  \textcolor{blue}{0.9942}    \\
                   & \multicolumn{2}{c}{STAA$_{soft}$}         &   \textcolor{blue}{46.59}   &   \textcolor{blue}{0.9942}   \\
                   & \multicolumn{2}{c}{STAA$_{quant}$}         &   \textcolor{red}{46.75}   &   \textcolor{red}{0.9944}   \\
                   & \multicolumn{2}{c}{STAA$_{ada}$}         &   45.32   &  0.9929    \\
                  \hline
2$\times t$/2$\times s$             & Nearest & Bicubic &  28.88    &  0.9073    \\
                  & \multicolumn{2}{c}{Gaussian}            & 37.44     &   0.9679   \\
                  & \multicolumn{2}{c}{STAA$_{no}$}         &   39.44   &     0.9775 \\
                   & \multicolumn{2}{c}{STAA$_{soft}$}         &   \textcolor{blue}{40.40}   &   \textcolor{red}{0.9812}   \\
                   & \multicolumn{2}{c}{STAA$_{quant}$}         &    \textcolor{red}{40.42}  &   \textcolor{blue}{0.9811}   \\
                    & \multicolumn{2}{c}{STAA$_{ada}$}         &  38.13    &   0.9720   \\
\hline
\end{tabular}
}
\end{table}

\noindent
\textbf{Constraints of the encoded frames.} The classical downsampling filters, \eg, nearest and Gaussian, can naturally guarantee that the downsampled frames resemble the input's appearance. However, in our auto-encoder framework, there is actually no guarantee that the encoded frames look like the original input. A straightforward way is to use the downsampled results from these classical methods as supervision, but it might impede the downsampler from learning the optimal spatio-temporal representation. So we turn to control the encoded frames implicitly by constraining the downsampling filter. We experiment three types of learned filters: (1) \textit{no}: no constraints added; (2) \textit{soft}: use the softmax to regularize the weight values; (3) \textit{quant}: add the differentiable quantization layer; (4)\textit{ada}: dynamically generate filters for each spatial location according to the input content (also with softmax). 

From the last four rows of Table~\ref{tab:abl_filter}, all STAA filters outperform the classical ones for reconstruction. The filter without any constraint is not necessary to be low-pass. Besides, it may cause color shifts in the encoded frames. Constraining the filter weights with softmax can alleviate color shifts and improve the reconstruction results due to anti-aliasing. Still, the moving regions are encoded as the color difference. Adding the quantization layer does not cause performance degradation, which validates the effectiveness of our differentiable implementation. Making the filter weights conditioned on the input content creates visually pleasing LR frames. However, the reconstruction performance degrades, probably because the changing weights of the downsampling filter confuse the upsampler. 

Comparing different downsampling settings, we observe that our STAA is more robust to temporal downsampling than previous methods. Specifically, the reconstruction quality is correlated to the logarithm of the percentage of pixels in the downsampling representation. More discussions are in our \textit{Appendix}.

\noindent
\textbf{Effectiveness of proposed modules.}
In Table~\ref{tab:abl_module}, we compare the video reconstruction results and the computational cost with different modules of the upsampler. We check the FLOPs per million pixels (MP) using the open-source tool fvcore~\cite{fvcore}. From the first row, we can observe that naive 3D convolution performs bad. Changing it to a more complex 3D residual-dense block (RDB) improves the performance by 2.68 dB, with a rapid increase of the computational cost. Although this network still cannot explicitly find the temporal correspondence, the deeper structure enlarges the perceiving area, thus enabling capturing dependencies with large displacement. In the third row, adopting deformable temporal modeling (DTM) shows a great performance improvement with relatively low computational cost, which validates the importance of aggregating the displaced information across space and time. Such spatio-temporal aggregated features can be effectively utilized by the 3D CNN, resulting in improved PSNR and SSIM results (see the last row).

\begin{table}[t]
\begin{center}
\caption{Ablation of modules in the upsampler.}
\label{tab:abl_module}
\resizebox{\linewidth}{!}{
\begin{tabular}{c|c|c|cccc}
\hline
Naive 3D Conv & 3D RDB & DTM & Params/M & GFlops/MP & PSNR  & SSIM   \\ \hline
\cmark           &     &        & 0.3    & 3.2     & 28.67 & 0.8536 \\
              &  \cmark    &      & 11.0     & 114.4   & 31.35 & 0.9016 \\
              &    &   \cmark     & 5.3      & 51.7    & 31.55 & 0.9032 \\
              & \cmark   & \cmark      & 16.0     & 164.7   & 32.00 & 0.9109 \\ 
              \hline
\end{tabular}
}
\end{center}
\vspace{-6mm}
\end{table}

\subsection{Applications}
The STAA downsampler can be used to reduce the resolution and frame rate of a video for efficient video transmission. The learned upsampler can also be applied to process natural videos with a simple modification. Besides, as discussed in Section~\ref{subsec: downsampler}, motion blur is a natural space-time low-pass filter. So our upsampler network can also reconstruct crisp clean frames from a blurry sequence. The training details of each application are in the \textit{supplementary material}.

\noindent
\textbf{Video Resampling.}
The space-time pixel-shuffle module makes it possible to change the frame rate with arbitrary ratios while keeping the motion patterns. Video frame interpolation methods can only synthesize new frames in-between the input frames, so the scale ratio can only be integers. Tools like ffmpeg~\cite{tomar2006converting} change frame rate by dropping or duplication at a certain interval, which changes the motion pattern of the original timestamps and cannot generate smooth results. Another option is to use frame blending to map the intermediate motion between keyframes while creating fuzzy and ghosting artifacts. Some softwares adopt optical flow warping, which can synthesize better results than the above two methods. Still, it cannot handle large motions or morph.

\begin{figure*}[tbp]
\captionsetup[subfigure]{labelformat=empty}
\begin{center}
\includegraphics[width=0.95\linewidth]{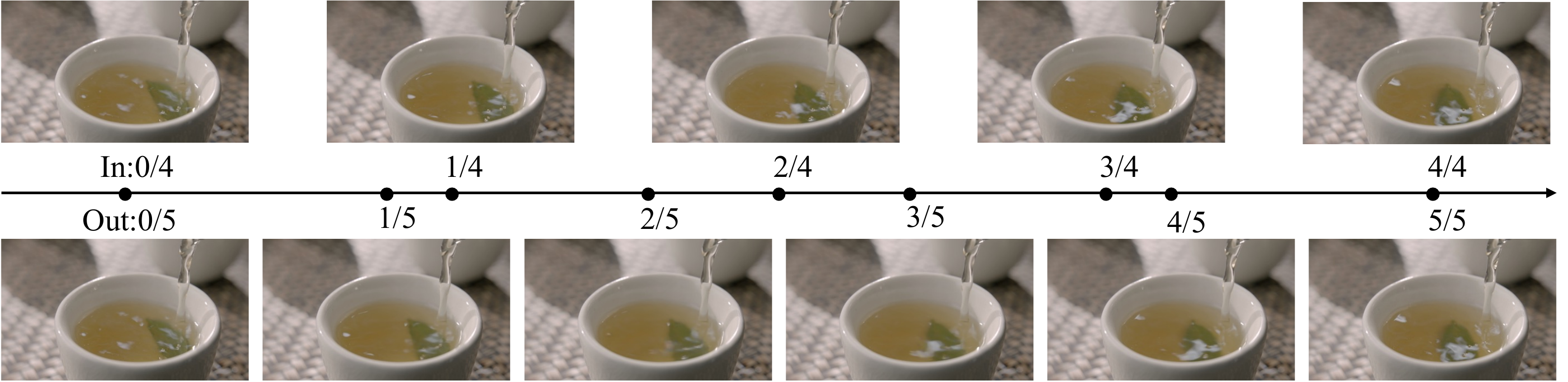}  
\end{center}  
\vspace{0mm}
\caption{Our methods enable smooth frame rate conversion with arbitrary rates, \eg{} 20 fps to 24 fps, which needs convert 5 frames into 6 frames with the same time length. The top row shows the input frames at 20 fps, and the bottom row shows the output sequence at 24 fps. We plot a timeline in the middle and mark the timestamp of each frame. The generated frames appear natural in motion transition, and have vivid textures, \eg water flow, reflection, and refraction.} 
\label{fig:24tp30fps}
\end{figure*}

Our upsampler can maintain the space-time patterns when upscaling the temporal dimension at any given ratio: we show an example of converting 20 fps to 24 fps (1.2$\times t$) in Fig.~\ref{fig:24tp30fps}, which does synthesize the correct motion at the non-existent time steps. The well-placed motions lead to smoother visual results. Our temporal modeling module can map long-range dependencies among the input frames, and together with the space-time convolutional layers, they can reconstruct sharp and crisp frames. To the best of our knowledge, this is the first deep-learning-based technique for arbitrary frame rate conversion.

\begin{figure*}[ht]
\captionsetup[subfigure]{labelformat=empty}
\begin{center}
  \begin{subfigure}[b]{\fivewidth\linewidth}
  \includegraphics[width=\linewidth]{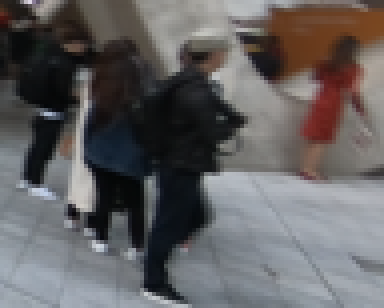}
  \end{subfigure}
  \begin{subfigure}[b]{\fivewidth\linewidth}
  \includegraphics[width=\linewidth]{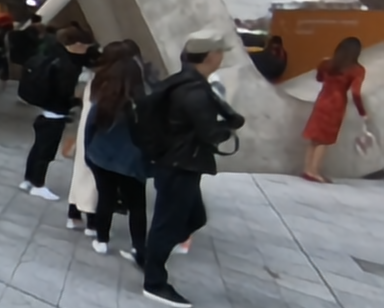}
  \end{subfigure}
  \begin{subfigure}[b]{\fivewidth\linewidth}
  \includegraphics[width=\linewidth]{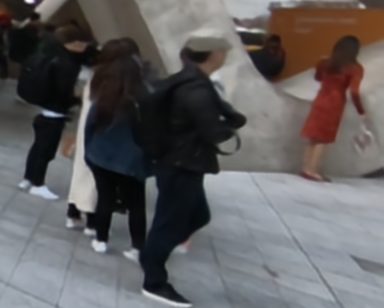}
  \end{subfigure}
  \begin{subfigure}[b]{\fivewidth\linewidth}
  \includegraphics[width=\linewidth]{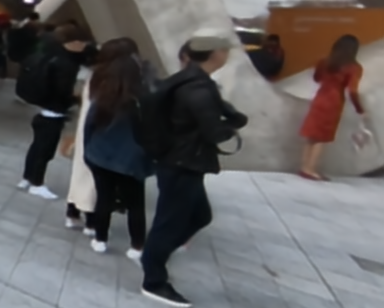}
  \end{subfigure}
  \begin{subfigure}[b]{\fivewidth\linewidth}
  \includegraphics[width=\linewidth]{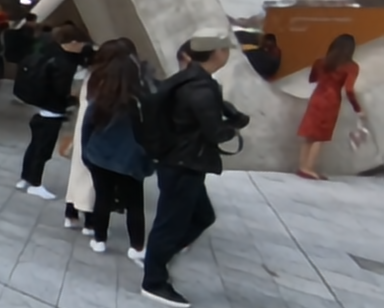}
  \end{subfigure}
  
  \begin{subfigure}[b]{\fivewidth\linewidth}
  \includegraphics[width=\linewidth]{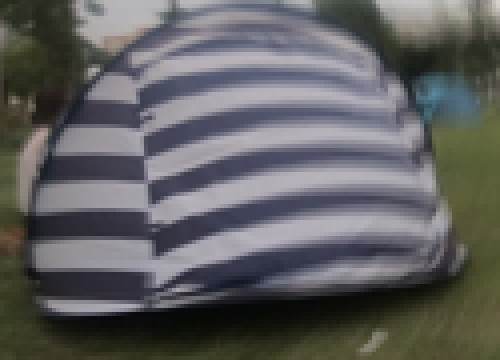}
  \subcaption{LR blurry input}
  \end{subfigure}
  \begin{subfigure}[b]{\fivewidth\linewidth}
  \includegraphics[width=\linewidth]{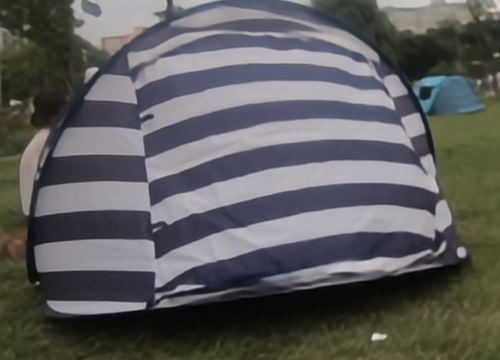}
  \subcaption{Output: 0}
  \end{subfigure}
  \begin{subfigure}[b]{\fivewidth\linewidth}
  \includegraphics[width=\linewidth]{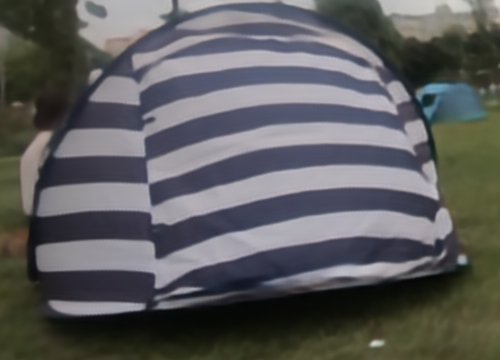}
  \subcaption{Output: 1}
  \end{subfigure}
  \begin{subfigure}[b]{\fivewidth\linewidth}
  \includegraphics[width=\linewidth]{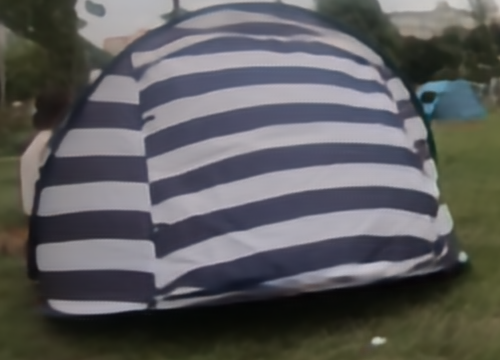}
  \subcaption{Output: 2}
  \end{subfigure}
  \begin{subfigure}[b]{\fivewidth\linewidth}
  \includegraphics[width=\linewidth]{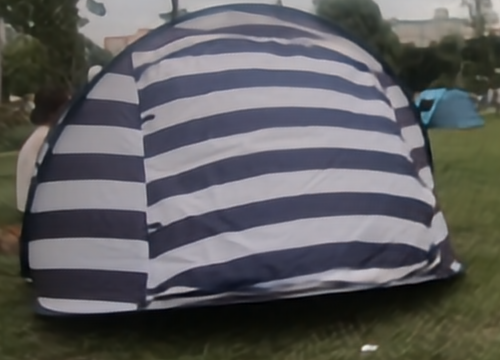}
  \subcaption{Output: 3}
  \end{subfigure}
\end{center}
\vspace{0mm}
\caption{Our upsampler can also be used for reconstructing sharp and crisp details from videos with motion-blur. The left column shows the overlaid two LR blurry inputs, and the right four columns are our reconstruction results with $2\times$ in time and $4\times$ in space, which recover shapes and textures from motion blur.}
\label{fig:bin}
\vspace{0mm}
\end{figure*}

\noindent
\textbf{Blurry Frame Reconstruction.}
As discussed in Sec.~\ref{subsec: downsampler}, motion blur is a temporal low-pass filter. It is a real-world case of our STAA filter: the temporal kernel size is the exposure time window, and the weights at each time step are equal. Hence, there is a good reason to believe that our designed upsampler can be applied on blurry frame reconstruction, which turns the low-resolution blurry sequence into a high frame-rate and high-resolution clean sequence. We trained our upsampler with a $4 \times s, 2 \times t$ upscale setting using the REDS-blur~\cite{Nah_2019_CVPR_Workshops_REDS} data. We show the restoration images in Fig.~\ref{fig:bin}. Even when the motion is rather large and the object texture is badly smeared, our upsampler does a good job in reconstructing the shape and structures at the correct timestep.

\noindent
\textbf{Efficient Video Storage and Transmission.}
Since the downsampler output is still in the same color space and data type (\eg 8-bit RGB) as the input, it can be processed by any existing encoding and compression algorithms for storage and transmission without extra elaborations. Especially, the downsampled frames still preserve the temporal connections implying their compatibility with video codecs.

\section{Conclusions}
\noindent
In this paper, we propose to learn a space-time downsampler and upsampler jointly to optimize the coding of the intermediate downsampled representations and ultimately boost video reconstruction performance. The downsampler includes a learned 3D low-pass filter for spatio-temporal anti-aliasing and a differentiable quantization layer to ensure the downsampled frames are encoded as standard 8-bit RGB images. For the upsampler, we propose the space-time pixel-shuffle to enable upscaling the $xyt$ volume at any given ratio. We further exploit the temporal correspondences between consecutive frames by explicit temporal modeling.
Due to the advantages of these designs, our framework outperforms state-of-the-art works in VSR and VFI by a large margin. Moreover, we demonstrate that our proposed upsampler can be used for highly accurate arbitrary frame-rate conversion, generating high-fidelity motion and visual details at the new timestamps for the first time. Our network can also be applied to blurry frame reconstruction and efficient video storage. We believe that our approach provides a new perspective on space-time video super-resolution tasks and has a broad potential to inspire novel methods for future works \eg quantization-aware image/video reconstruction, restoration-oriented video compression, and hardware applications such as coded exposure and optical anti-aliasing filter.

\section*{Acknowledgements}
\noindent We thank Hai Wang for the quantitative results of previous SOTA methods, Salma Abdel Magid for feedback on the manuscript, Chakravarty Reddy Alla Chaitanya and Siddharth Bhargav for discussions.

{\small
\bibliographystyle{ieee_fullname}
\bibliography{egbib}
}

\clearpage

\begin{appendices}

In this appendix, we first discuss the difference between video frame interpolation and temporal upscaling in Sec.~\ref{sec:diff}. Then, we provide more module design details in Sec.~\ref{sec:module_design_details}. Besides, we provide the background knowledge of space-time Fourier analysis in Sec.~\ref{sec:back_fourier} to help the readers understand the relevant part in the main paper. Sec.~\ref{sec:supp_exp} includes more comparisons with the state-of-the-art (SOTA) methods and the results of integrating other network structures in the Space-Time Anti-Aliasing (STAA) framework. Lastly, we show the training setting and more results of the extensive applications in Sec.~\ref{sec:ext_app}.

\section{Difference between Video Frame Interpolation and Temporal Upscaling}
\label{sec:diff}
In this section, we explain the difference between the video frame interpolation (VFI) task and the temporal upscaling proposed in this paper. 

\begin{figure*}[hbpt]
\captionsetup[subfigure]{labelformat=empty}
\begin{center}
  \begin{subfigure}[b]{\twowidth\linewidth}
  \includegraphics[width=\linewidth]{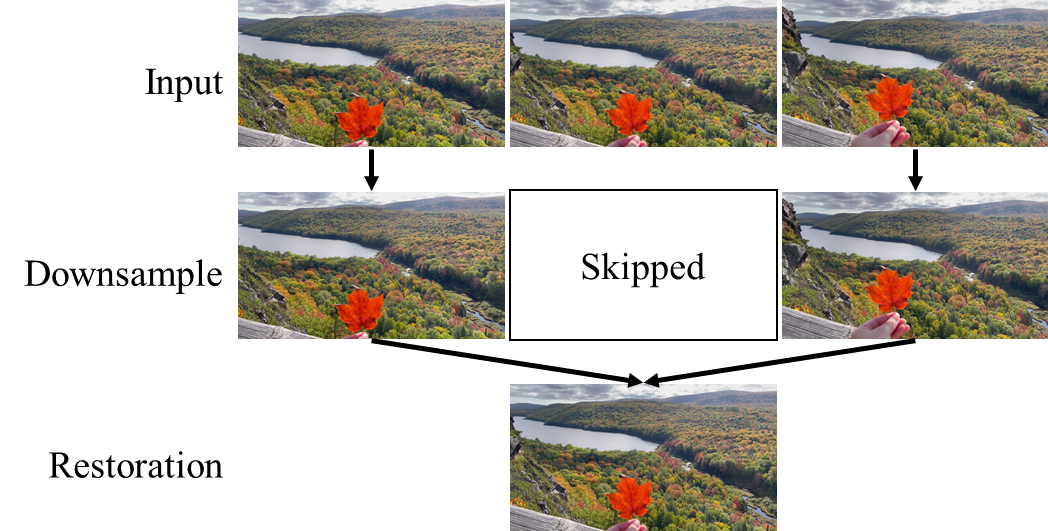}
  \subcaption{(a) Video Frame Interpolation}
  \end{subfigure}
 \begin{subfigure}[b]{\twowidth\linewidth}
 \includegraphics[width=\linewidth]{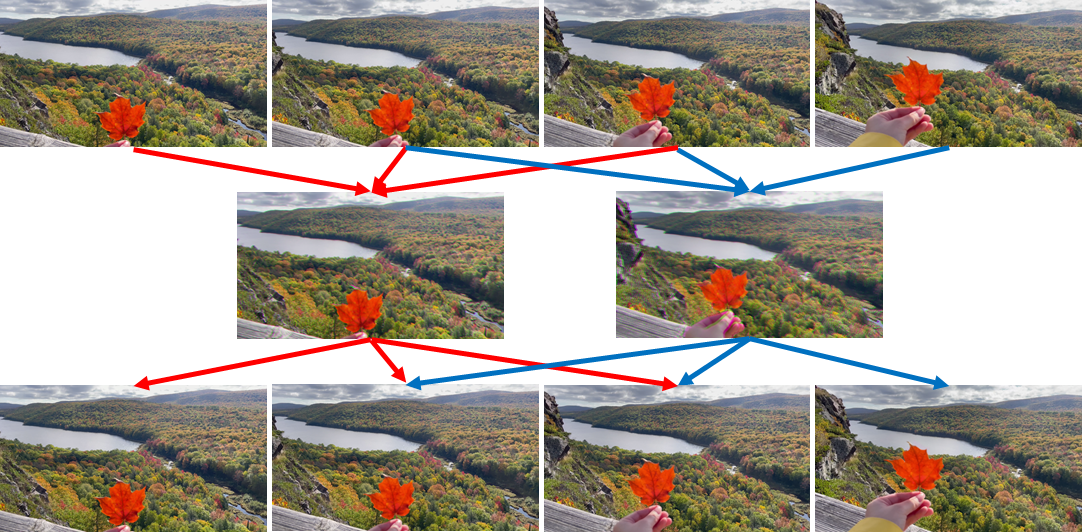}
  \subcaption{(b) Temporal Upscaling}
  \end{subfigure}
\end{center}
\caption{\textit{Illustration of the difference between video frame interpolation and temporal upscaling.} In VFI settings, each downsampled frame corresponds to only a single frame of the original input. While for temporal upscaling, we allow the downsampled frame to assimilate information from the adjacent frames of the original input. Thus, we could reconstruct a full sequence from it.}
\label{fig:diff_illust}
\end{figure*}

Fig.~\ref{fig:diff_illust} shows the illustration of these two different settings: given an input sequence of frames $I_1, I_2, I_3, I_4, \ldots$, VFI adopts nearest-neighbor downsampling to acquire the downsampled sequence $I_1, I_2, \ldots$, as shown in (a). The acquired sequence explicitly corresponds to the timestep in the original input. In the restoration step, it needs to synthesize the missing frames from neighbors in the downsampled sequence: \eg
\begin{equation}
    I_1, I_3 \rightarrow \widetilde{I}_2.
\end{equation} 
Thus, in the final sequence with a total length of $2n-1$, $n-1$ frames are synthesized and $n$ frames are directly copied from the original input. So VFI methods cannot perfectly reconstruct a sequence with an even number of frames.

Temporal upscaling is more like spatial-upscaling: given an input sequence $I_1, I_2, I_3, I_4, \ldots$, the anti-aliasing filter blends the information from nearby frames, thus each frame of the downsampled sequence should contain information from multiple frames of the original input. Thus, it does not explicitly correspond to the timestep in the original input. So we annotate the downsampled frames as $I_{123}, I_{234}, \ldots$. Since this downsampled representation contains the motion information at each timestep of the original sequence, we should be able to reconstruct them with proper design. For the restoration output, all frames in the sequence are synthesized: 
\begin{equation}
I_{123}, I_{234} \rightarrow \widetilde{I}_1, \widetilde{I}_2, \widetilde{I}_3, \widetilde{I}_4.    
\end{equation}
The temporal upscaling method can reconstruct a sequence with either an even or odd number of frames.

Due to the above differences, it is unfair to directly compare the output between these two methods. So in this paper, we only compare the frames that are synthesized by both methods (\eg in the above example, $\widetilde{I}_2, \widetilde{I}_4, \ldots$) to measure the reconstruction capability of the network. For the full sequence, we compare the temporal profile to evaluate the reconstructed motion patterns.

\section{Module Design Details}
\label{sec:module_design_details}
In this section, we provide more details about the two modules of the downsampler: the space-time anti-aliasing filter and the differentiable quantization layer, and the two modules of the upsampler: the deformable temporal modeling (DTM) and the residual dense block with 3D convolutions.

\subsection{Space-Time Anti-Aliasing Downsampler}
\begin{figure}[tbp]
\centering
\includegraphics[width=\linewidth]{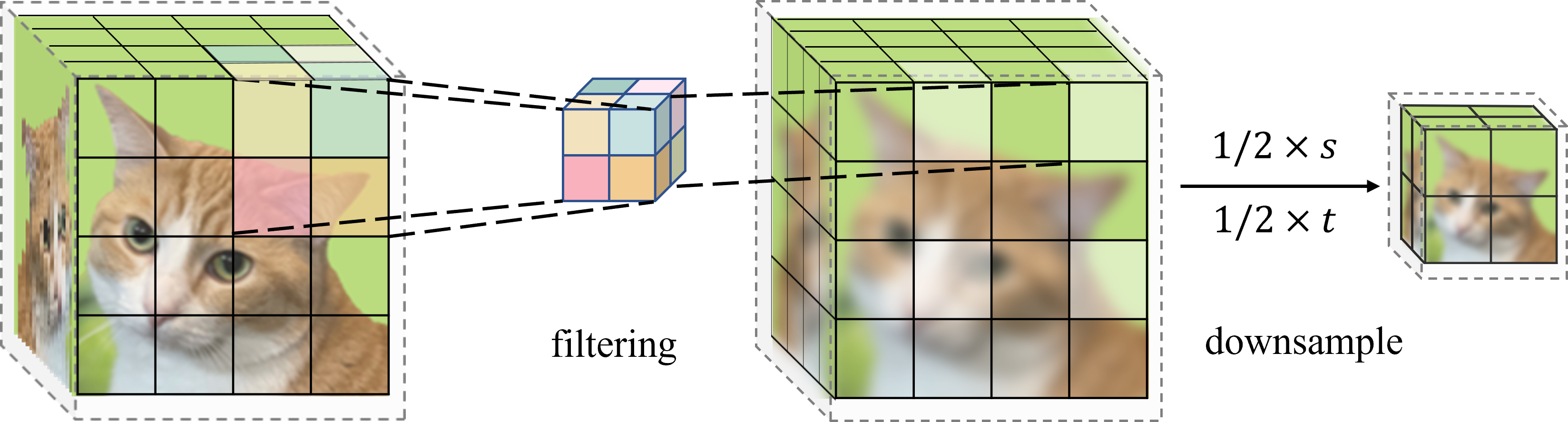}
\caption{Illustration of our learned space-time anti-aliasing downsampler.}
\label{fig:teaser_sta_operation}
\vspace{-3mm}
\end{figure}

Fig.~\ref{fig:teaser_sta_operation} illustrates the two operations in the downsampler: filtering with the learned 3D space-time filter and downsampling with stride. In our implementation, the above two steps are simplified into a convolution: the convolution kernel size is the window size in space/time; the convolution weights are learnable with custom constraints; the downsampling in each dimension is controlled by stride.

\subsection{Differentiable Quantization Layer}

The direct output of our downsampler is a floating-point tensor, while in practical applications, images are usually encoded as 8-bit RGB (uint8). To make our downsampled frames compatible with popular image data storage and transmission pipelines, we propose a quantization layer and include it in our end-to-end training process. 

Directly casting the tensor type to uint8 does not work: such operations are not differentiable and thus cannot be used to train our network in an end-to-end manner. Consequently, if we directly send a uint8 input to the upsampler, there is going to be a performance drop due to the precision gap between the float and the uint8. So we split the quantization for tensors into two steps: 

(1) \textbf{Clipping}: limit the values to a certain range (\eg [0, 255] for uint8) to avoid blown-out colors. There is a serious problem: the derivative out of the $[min, max]$ value is masked to be 0, which makes the training at early iterations unstable, as reported in~\cite{kim2018task}. To solve this problem, we modify the clipping function $Q(\cdot)$ into:
\begin{equation}
\begin{split}
    \hat{x} & = x + \text{ReLU}(min-x), \\
    Q(x) & = \hat{x} - \text{ReLU}(\hat{x}-max),
\end{split}
\end{equation}
where $x$ is our input value, $[min, max]$ denotes the clipping range, and $Q(x)$ is the clipped output. The derivative of the clipping function becomes:
\begin{equation}
    \dfrac{d Q(x)}{dx} =  \begin{cases}
      2, & \text{if $x \leq min$} \\
      1, & \text{if $x \in (min, max)$}\\
      2, & \text{if $x \geq max$}
      \end{cases}
\end{equation}

This modification does not influence clipping results but makes the derivative non-zero everywhere.

(2) \textbf{Round}: round the values to the closest integer. The gradient of this function is zero almost everywhere, which is bad for optimization. To avoid this issue, we override the gradient map to 1 during backward propagation.

\begin{figure}[t]
\captionsetup[subfigure]{labelformat=empty}
\begin{center}
  \begin{subfigure}[b]{\twowidth\linewidth}
  \includegraphics[width=\linewidth]{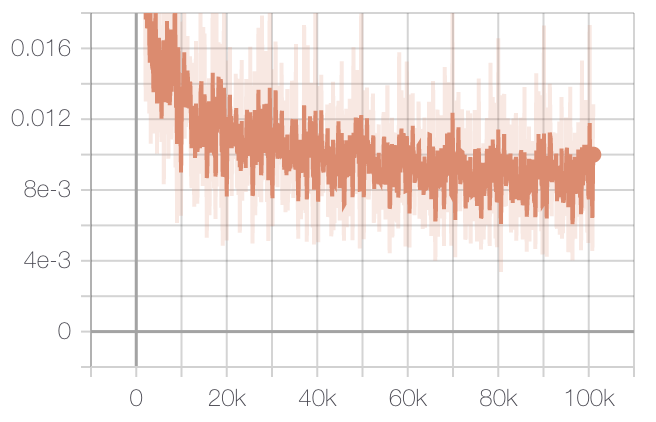}
  \subcaption{Training Loss}
  \end{subfigure}
 \begin{subfigure}[b]{\twowidth\linewidth}
 \includegraphics[width=\linewidth]{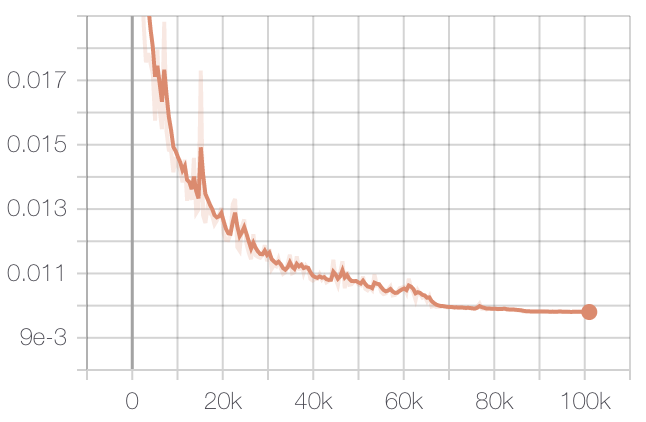}
  \subcaption{Validation Loss}
  \end{subfigure}
\end{center}
\caption{\textit{Plot of our training progress}. With the aid of the differentiable quantization layer, our network's training is stable from scratch.}
\label{fig:train_curve}
\vspace{-3mm}
\end{figure}

As a result, the quantization operation's derivative is non-zero everywhere, thus enabling end-to-end training. Fig.~\ref{fig:train_curve} shows the training and validation loss of our framework. It shows our proposed method can be trained stably from scratch. 

\subsection{Deformable Temporal Modeling (DTM)}
\begin{figure}[tbp]
\centering
\includegraphics[width=\linewidth]{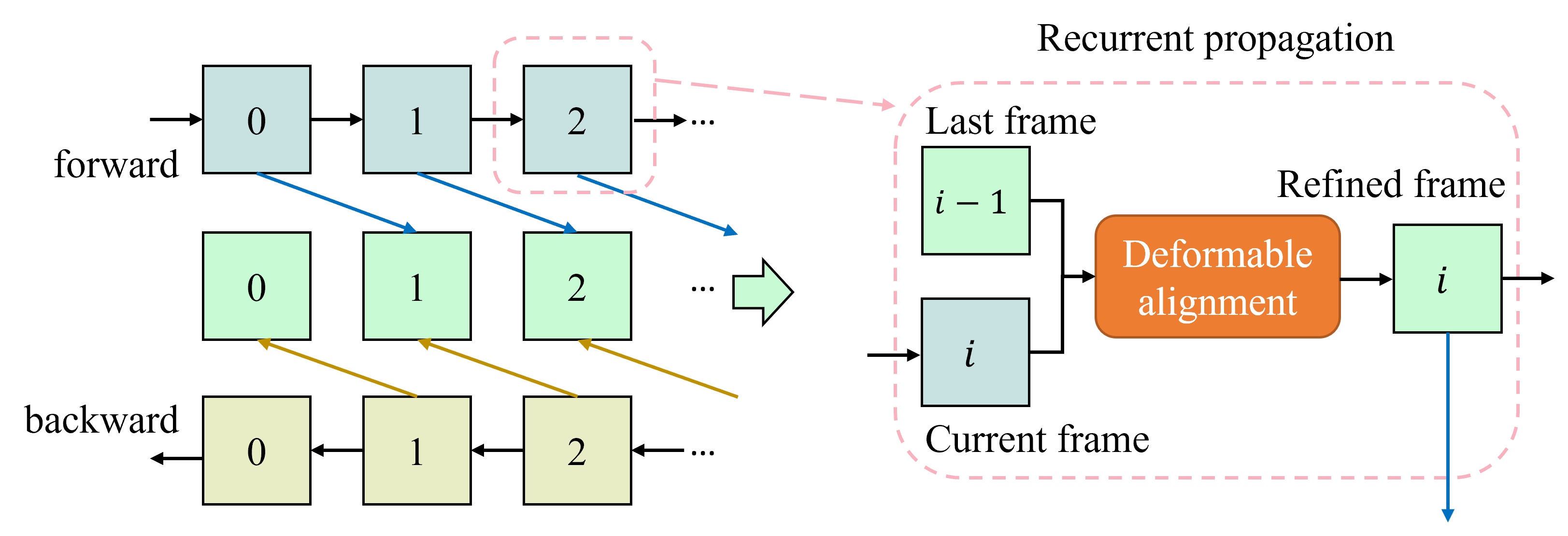}
\caption{\textit{Illustration of the deformable temporal modeling network.} The temporal propagation is implemented in a recurrent manner with forward and backward directions. At each step, we model the temporal correspondence between the last refined frame and the current frame with deformable alignment and output the refined current frame.}
\label{fig:dtm}
\end{figure}

Here we describe the detailed structure of the deformable temporal module in Fig.~\ref{fig:dtm}: as illustrated in the main paper, the temporal correspondences are propagated in both forward and backward directions. At each step $i$, it takes a current feature $f_i$ as input and refine it with the last step's output $f'_{i-1}$ with deformable alignment function $T(\cdot)$:
\begin{equation}
    f'_i = T(f_i, f'_{i-1}).
\end{equation}

The deformable alignment function $T(\cdot)$ is basically a deformable convolution, which takes an input feature map $f'_{i-1}$ and an offset map $\Delta p_i$:
\begin{equation}
    T(f_i, f'_{i-1}) = DConv(f'_{i-1}, \Delta p_i),
\end{equation}
where the offset $\Delta p_i$ is estimated from the refined feature map from the last step and the current step with convolutional function $g(\cdot)$:
\begin{equation}
    \Delta p_i = g(f'_{i-1}, f_i).
\end{equation}

In this way, the previous feature maps are aligned with the current one, which enhances our network's capability of handling motions. Then we aggregate the current feature $f_i$ and the aligned feature $f'_i$ to get the refined feature: $r(f_i) = c(f_i, f'_i)$ with the ConvLSTM $c$. After bi-directional propagation, for each timestep $i$, there will be two refined features from each direction. The aggregated output can be acquired by a blending function:
\begin{equation}
    DTM(f_i) = w_f*r_f(f_i) + w_b r_b(f_i),
\end{equation}
where $w_f, w_b$ are conv$1\times 1$ kernels and $*$ denotes the convolutional operator. In this way, we get a well-aligned sequence as output.

\subsection{Residual Dense Block with 3D Convolutions}
\begin{figure}[tbp]
\centering
\includegraphics[width=\linewidth]{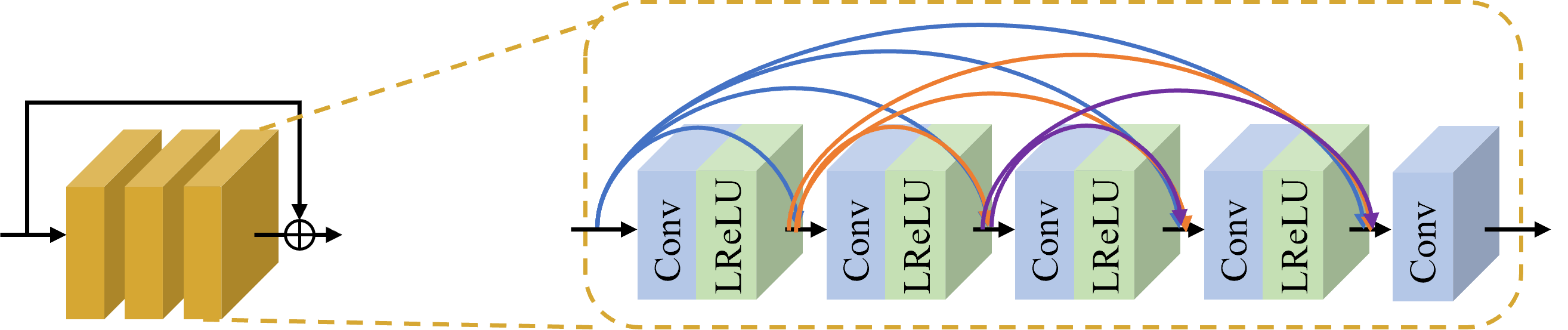}
\caption{Illustration of the residual dense block with 3D convolution.}
\label{fig:rdb}
\end{figure}

As the major building block of the reconstruction trunk in our upsampler, the 3D convolutions are organized exactly the same way as the residual dense block in ~\cite{zhang2018residual,wang2018esrgan} while replacing the 2D convolutions with 3D, as shown in Fig.~\ref{fig:rdb}. The skip-connection of the residual block is with a residual scaling parameter $\beta$; in the dense block, features of different hierarchical levels are concatenated.  

\section{Background of Space-Time Fourier Analysis}
\label{sec:back_fourier}
Given a video signal $f(x,y,t)$, we can analyze its space-time frequency characteristics through Fourier transform ($\mathcal{F}(\cdot)$):
\begin{equation} \label{eq:f}
    F(\Omega_x, \Omega_y, \Omega_t) = \mathcal{F}_{x,y,t}(f(x,y,t)),
\end{equation}
where $\Omega$ denotes the frequency for each dimension, and $F$ is the Fourier transform of $f$. For the uniform-velocity case as discussed in our main paper, we can denote the constant velocities as $v_x$ and $v_y$ for each direction. Thus, the object signal is translated through time by:
\begin{equation}
    f(x,y,t) = g(x-v_x t, y-v_y t),
\end{equation}
where $g(x,y)$ is the signal of the 2D object. We denote its Fourier transform as $G$. Correspondingly, the Fourier transform in Eq.~\ref{eq:f} becomes:
\begin{equation} \label{eq:f_long}
\begin{split}
    F(\Omega_x, \Omega_y, \Omega_t) & = \mathcal{F}_{x,y,t}[g(x-v_x t, y-v_y t)] \\
     & = G(\Omega_x \Omega_y)\int e^{-2\pi i t(\Omega_x v_x + \Omega_y v_y + \Omega_t)} dt \\
     & = G(\Omega_x \Omega_y)\delta(\Omega_x v_x + \Omega_y v_y + \Omega_t).
\end{split}
\end{equation}

This explains why moving the 2D object would result in a shearing in the frequency domain: theoretically, all the non-zero frequency components should lie on the plane $\Omega_x v_x + \Omega_y v_y + \Omega_t$ in the 3D Fourier space, which reflects the coupling of space and time dimensions. However, our digital image is discrete -- our simulated temporal profile has jigsaw-like boundaries between the moving object and the background due to the motion being at the pixel level. As a result, the transformed figure has two small sub-bands, as shown in the main paper. For visualization purposes, our analysis in the main paper is on the $xt$ 2D signals, thus Eq.~\ref{eq:f_long} can be written as:
\begin{equation}
    F(\Omega_x, \Omega_t) = G(\Omega_x)\delta(\Omega_x v_x + \Omega_t),
\end{equation}
which shows the spectrum is limited by the delta function to a single line $\Omega_x v_x + \Omega_t=0$. The velocity $v_x$ decides the slope of this line. For the $xt$ 2D signal, convolving with a filter $h(\cdot)$ equals to a multiplication with $H(\cdot)$ in the Fourier domain:
\begin{equation}
    H_F(\Omega_x, \Omega_t) = G(\Omega_x)\delta(\Omega_x v_x + \Omega_t)H(\Omega_x, \Omega_t).
\end{equation}

This explains the effect of different filters: Gaussian filter's Fourier transform is still a Gaussian function, which gradually attenuates the spatio-temporal high frequency; box filter (motion blur)'s Fourier transform is a sinc function, which attenuates certain frequency components. 

\section{More Experiments}
\label{sec:supp_exp}
\subsection{More Comparisons with State-of-the-Art Methods
}
\begin{figure*}[t]
\captionsetup[subfigure]{labelformat=empty}
\begin{center}

  \begin{subfigure}[b]{\fivewidth\linewidth}
  \includegraphics[width=\linewidth]{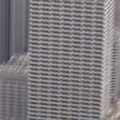}
  \end{subfigure}
  \begin{subfigure}[b]{\fivewidth\linewidth}
  \includegraphics[width=\linewidth]{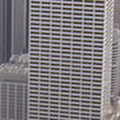}
  \end{subfigure}
  \begin{subfigure}[b]{\fivewidth\linewidth}
  \includegraphics[width=\linewidth]{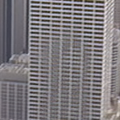}
  \end{subfigure}
  \begin{subfigure}[b]{\fivewidth\linewidth}
  \includegraphics[width=\linewidth]{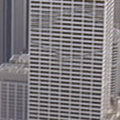}
  \end{subfigure}
  \begin{subfigure}[b]{\fivewidth\linewidth}
  \includegraphics[width=\linewidth]{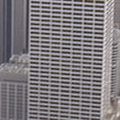}
  \end{subfigure}

  \begin{subfigure}[b]{\fivewidth\linewidth}
  \includegraphics[width=\linewidth]{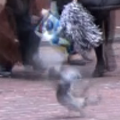}
  \end{subfigure}
  \begin{subfigure}[b]{\fivewidth\linewidth}
  \includegraphics[width=\linewidth]{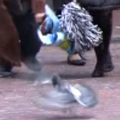}
  \end{subfigure}
  \begin{subfigure}[b]{\fivewidth\linewidth}
  \includegraphics[width=\linewidth]{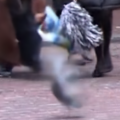}
  \end{subfigure}
  \begin{subfigure}[b]{\fivewidth\linewidth}
  \includegraphics[width=\linewidth]{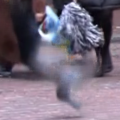}
  \end{subfigure}
  \begin{subfigure}[b]{\fivewidth\linewidth}
  \includegraphics[width=\linewidth]{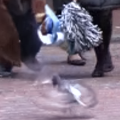}
  \end{subfigure}

  \begin{subfigure}[b]{\fivewidth\linewidth}
  \includegraphics[width=\linewidth]{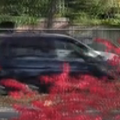}
  \subcaption{Input-Overlayed}
  \end{subfigure}
  \begin{subfigure}[b]{\fivewidth\linewidth}
  \includegraphics[width=\linewidth]{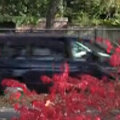}
  \subcaption{GT}
  \end{subfigure}
  \begin{subfigure}[b]{\fivewidth\linewidth}
  \includegraphics[width=\linewidth]{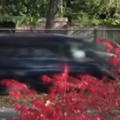}
  \subcaption{FLAVR~\cite{kalluri2020flavr}}
  \end{subfigure}
  \begin{subfigure}[b]{\fivewidth\linewidth}
  \includegraphics[width=\linewidth]{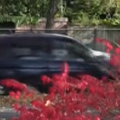}
  \subcaption{XVFI~\cite{sim2021xvfi}}
  \end{subfigure}
  \begin{subfigure}[b]{\fivewidth\linewidth}
  \includegraphics[width=\linewidth]{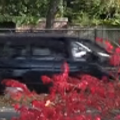}
  \subcaption{\textbf{Ours}}
  \end{subfigure}
\end{center} 
\vspace{-3mm}
\caption{Qualitative results of $1\times s, 2\times t$ upscaling on Vid4 dataset. Compared with previous video frame interpolation methods, the reconstruction results of our proposed STAA framework have better textures while preserving the motion patterns due to the co-design of downsampler and upsampler.} 
\label{fig:comp_sota_s1t2}
\vspace{-3mm}
\end{figure*}

\begin{figure*}[tbp]
\captionsetup[subfigure]{labelformat=empty}
\begin{center}
  \begin{subfigure}[b]{\fourwidth\linewidth}
  \includegraphics[width=\linewidth]{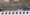}
  \subcaption{Input-Overlayed}
  \end{subfigure}
  \begin{subfigure}[b]{\fourwidth\linewidth}
  \includegraphics[width=\linewidth]{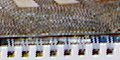}
  \subcaption{GT}
  \end{subfigure}
  \begin{subfigure}[b]{\fourwidth\linewidth}
  \includegraphics[width=\linewidth]{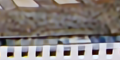}
  \subcaption{SepConv~\cite{niklaus2017adsconv}+EDVR~\cite{wang2019edvr}}
  \end{subfigure}
  \begin{subfigure}[b]{\fourwidth\linewidth}
  \includegraphics[width=\linewidth]{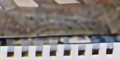}
  \subcaption{DAIN~\cite{bao2019depth}+EDVR~\cite{wang2019edvr}}
  \end{subfigure}
  
  \begin{subfigure}[b]{\fourwidth\linewidth}
  \includegraphics[width=\linewidth]{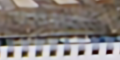}
  \captionsetup[sub]{font={small}}
  \subcaption{FLAVR~\cite{kalluri2020flavr}+BasicVSR++~\cite{chan2021basicvsr++}}
  \end{subfigure}
  \begin{subfigure}[b]{\fourwidth\linewidth}
  \includegraphics[width=\linewidth]{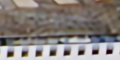}
  \subcaption{XVFI~\cite{sim2021xvfi}+BasicVSR++~\cite{chan2021basicvsr++}}
  \end{subfigure}
  \begin{subfigure}[b]{\fourwidth\linewidth}
  \includegraphics[width=\linewidth]{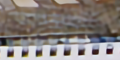}
  \subcaption{ZSM~\cite{xiang2020zooming}}
  \end{subfigure}  
  \begin{subfigure}[b]{\fourwidth\linewidth}
  \includegraphics[width=\linewidth]{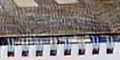}
  \subcaption{\textbf{Ours}}
  \end{subfigure}

  

  \begin{subfigure}[b]{\fourwidth\linewidth}
  \includegraphics[width=\linewidth]{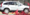}
  \subcaption{Input-Overlayed}
  \end{subfigure}
  \begin{subfigure}[b]{\fourwidth\linewidth}
  \includegraphics[width=\linewidth]{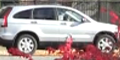}
  \subcaption{GT}
  \end{subfigure}
  \begin{subfigure}[b]{\fourwidth\linewidth}
  \includegraphics[width=\linewidth]{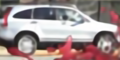}
  \subcaption{SepConv~\cite{niklaus2017adsconv}+EDVR~\cite{wang2019edvr}}
  \end{subfigure}
  \begin{subfigure}[b]{\fourwidth\linewidth}
  \includegraphics[width=\linewidth]{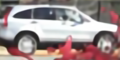}
  \subcaption{DAIN~\cite{bao2019depth}+EDVR~\cite{wang2019edvr}}
  \end{subfigure}
  
  \begin{subfigure}[b]{\fourwidth\linewidth}
  \includegraphics[width=\linewidth]{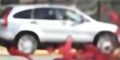}
  \captionsetup[sub]{font={small}}
  \subcaption{FLAVR~\cite{kalluri2020flavr}+BasicVSR++~\cite{chan2021basicvsr++}}
  \end{subfigure}
  \begin{subfigure}[b]{\fourwidth\linewidth}
  \includegraphics[width=\linewidth]{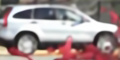}
  \subcaption{XVFI~\cite{sim2021xvfi}+BasicVSR++~\cite{chan2021basicvsr++}}
  \end{subfigure}
  \begin{subfigure}[b]{\fourwidth\linewidth}
  \includegraphics[width=\linewidth]{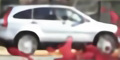}
  \subcaption{ZSM~\cite{xiang2020zooming}}
  \end{subfigure}  
  \begin{subfigure}[b]{\fourwidth\linewidth}
  \includegraphics[width=\linewidth]{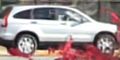}
  \subcaption{\textbf{Ours}}
  \end{subfigure}

  \begin{subfigure}[b]{\fourwidth\linewidth}
  \includegraphics[width=\linewidth]{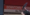}
  \subcaption{Input-Overlayed}
  \end{subfigure}
  \begin{subfigure}[b]{\fourwidth\linewidth}
  \includegraphics[width=\linewidth]{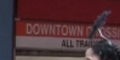}
  \subcaption{GT}
  \end{subfigure}
  \begin{subfigure}[b]{\fourwidth\linewidth}
  \includegraphics[width=\linewidth]{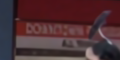}
  \subcaption{SepConv~\cite{niklaus2017adsconv}+EDVR~\cite{wang2019edvr}}
  \end{subfigure}
  \begin{subfigure}[b]{\fourwidth\linewidth}
  \includegraphics[width=\linewidth]{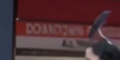}
  \subcaption{DAIN~\cite{bao2019depth}+EDVR~\cite{wang2019edvr}}
  \end{subfigure}
  
  \begin{subfigure}[b]{\fourwidth\linewidth}
  \includegraphics[width=\linewidth]{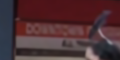}
  \captionsetup[sub]{font={small}}
  \subcaption{FLAVR~\cite{kalluri2020flavr}+BasicVSR++~\cite{chan2021basicvsr++}}
  \end{subfigure}
  \begin{subfigure}[b]{\fourwidth\linewidth}
  \includegraphics[width=\linewidth]{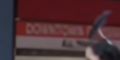}
  \subcaption{XVFI~\cite{sim2021xvfi}+BasicVSR++~\cite{chan2021basicvsr++}}
  \end{subfigure}
  \begin{subfigure}[b]{\fourwidth\linewidth}
  \includegraphics[width=\linewidth]{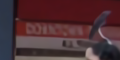}
  \subcaption{ZSM~\cite{xiang2020zooming}}
  \end{subfigure}  
  \begin{subfigure}[b]{\fourwidth\linewidth}
  \includegraphics[width=\linewidth]{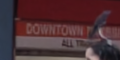}
  \subcaption{\textbf{Ours}}
  \end{subfigure}
  
\end{center} 
\vspace{-3mm}
\caption{Qualitative results of $4\times s, 2\times t$ upscaling on Vid4 dataset. Compared with previous two-stage and one-stage space-time super-resolution methods, the reconstruction results of our proposed STAA framework have much better textures while preserving the motion patterns due to the co-design of downsampler and upsampler.} 
\label{fig:comp_sota_s4t2}
\vspace{-3mm}
\end{figure*}

In this section, we first show more visual results that compare the previous reconstruction methods with our joint-learned STAA in Fig.~\ref{fig:comp_sota_s1t2} and Fig.~\ref{fig:comp_sota_s4t2}. 

Fig.~\ref{fig:comp_sota_s1t2} shows the temporal upscaling results ($1\times s, 2\times t$). Compared with previous VFI methods, the reconstruction results of our proposed STAA framework have more vivid spatial textures while preserving the correct motion patterns. Because our STAA downsampler can encode the motion of each frame in the original input sequence, our reconstruction result maintains the fast motions that are lost in the previous nearest-neighbor sampling (see the pigeon in the second row).

Fig.~\ref{fig:comp_sota_s4t2} shows the space-time upscaling results ($4\times s, 2\times t$). We choose SepConv~\cite{niklaus2017adsconv}, DAIN~\cite{bao2019depth}, FLAVR~\cite{kalluri2020flavr} and XVFI~\cite{sim2021xvfi} as VFI methods, and EDVR~\cite{wang2019edvr}, BasicVSR++~\cite{chan2021basicvsr++} as VSR methods, ZSM~\cite{xiang2020zooming,xiang2021zooming} as STVSR methods for comparison. Benefiting from the joint encoding of space-time dimensions, our reconstruction result has much richer spatial details even without the aid of perceptual or adversarial loss, which is almost impossible to be restored by previous methods. These results demonstrate the great advantage of co-designing the downsampler with the upsampler, which could be a potential direction for the video reconstruction task.

We also compare the computational efficiency of different space-time video upscaling methods by computing the FLOPs per million pixels (MP) using the open-source tool fvcore~\cite{fvcore}, as shown in Table~\ref{tab:flops}. Compared with the two-stage methods with cascaded VFI and VSR networks, the one-stage space-time super-resolution network is more efficient. Our proposed upsampler requires the least computation cost among all compared methods.

\begin{table}[t]
\caption{Comparison of computational costs among cascaded VFI and VSR methods, STVSR methods, and our method.}
\label{tab:flops}
\begin{center}
\begin{tabular}{cc|c}
\hline
\multicolumn{2}{c|}{Method} & \multirow{2}{*}{GFLOPs/MP} \\ 
VFI           & VSR        &                           \\ \hline
FLAVR~\cite{kalluri2020flavr}         &   BasicVSR++~\cite{chan2021basicvsr++}         &       185.34+140.29                \\
XVFI~\cite{sim2021xvfi}          &   BasicVSR++~\cite{chan2021basicvsr++}         & 676.65+140.29                       \\
\multicolumn{2}{c|}{ZSM~\cite{xiang2020zooming}}   & 198.51             \\
\multicolumn{2}{c|}{Ours}   & 163.98          \\
\hline
\end{tabular}
\end{center}
\vspace{-3mm}
\end{table}

\subsection{Our Framework Can be Generalized to More Networks}
To compare the influence of the joint-learning mechanism of our framework, we switch the upsampler to an STVSR network ZSM~\footnote{Note that the reconstructed result is not comparable with our video upscaling setting of $4\times s, 2\times t$, since our network decodes 8 frames from 4 LR inputs, which is more challenging than decoding 7 out of 4.}, which converts 4 LR frames to 7 HR frames. In the updated STAA-ZSM setup, the 4 LR frames are generated by our STAA downsampler and then sent to the ZSM for reconstruction. Compared with the original ZSM in Table~\ref{tab:sta_zsm}, we can observe that the STAA downsampled representation improves the PSNR by 2.14 dB and SSIM by 0.0348. These results demonstrate that the improvement brought by the joint learning framework can be generalized to other networks.

\begin{table}[t]
\caption{Using STAA filter for space-time super-resolution.}
\label{tab:sta_zsm}
\begin{center}
\begin{tabular}{c|cc}
\hline
 \multirow{2}{*}{Method} & \multicolumn{2}{c}{Vimeo-90k} \\ 
          & PSNR        &     SSIM                      \\ \hline
ZSM~\cite{xiang2020zooming} & 33.48 & 0.9178 \\
STAA-ZSM & \textbf{35.62} & \textbf{0.9526} \\ \hline
\end{tabular}
\end{center}
\vspace{-3mm}
\end{table}

To evaluate the effectiveness of our upsampler design, we train another framework under the setting $1\times s, 2\times t$ by substituting the upsampler with FLAVR~\cite{kalluri2020flavr}. We modify the input and output numbers of the FLAVR to make it compatible with the temporal upscaling ratio. From Table~\ref{tab:sta_flavr}, we can see that our proposed upsampler outperforms FLAVR by a large margin. Considering that both networks adopt 3D convolution as the basic building block, we believe that such improvement should attribute to the deformable temporal modeling. 

\begin{table}[tbp]
\caption{Performance comparison for temporal upscaling: $2\times t$.}
\label{tab:sta_flavr}
\begin{center}
\begin{tabular}{c|ccc}
\hline
 \multirow{2}{*}{Method} & \multicolumn{2}{c}{Vimeo-90k} & \multirow{2}{*}{Params(M)} \\ 
          & PSNR        &     SSIM                      \\ \hline
STAA-FLAVR~\cite{kalluri2020flavr} & 44.72 &	0.9915 & 42.1 \\
STAA-Ours & \textbf{46.59} &	\textbf{0.9942} & 15.9 \\ \hline
\end{tabular}
\end{center}
\end{table}

\section{Extensive Applications}
\label{sec:ext_app}
In this section, we provide the training details for each application, and show more results and analysis. Besides, we also discuss the trade-off between the data storage and the reconstruction performance.

\subsection{Frame Rate Conversion}
\noindent
\textbf{Training Details.}
We adopt REDS~\cite{Nah_2019_CVPR_Workshops_REDS} as our training set: it contains 270 videos of dynamic scenes at $720\times 1280$ resolution, in which 240 videos are split as the training set and 30 videos as the validation set. In this application, we train our upsampler network with the training set. During the training, we take the 5-frame sequence as input and the 6-frame sequence as the supervision for output. 

\begin{figure*}[tbp]
\captionsetup[subfigure]{labelformat=empty}
\begin{center}
  \begin{subfigure}[b]{\fivewidth\linewidth}
  \includegraphics[width=\linewidth]{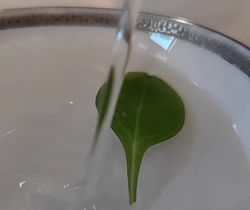}
  \subcaption{GT}
  \end{subfigure}
  \begin{subfigure}[b]{\fivewidth\linewidth}
  \includegraphics[width=\linewidth]{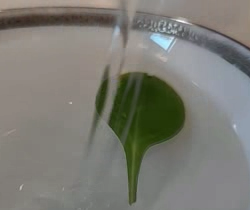}
  \subcaption{premiere-blend}
  \end{subfigure}
  \begin{subfigure}[b]{\fivewidth\linewidth}
  \includegraphics[width=\linewidth]{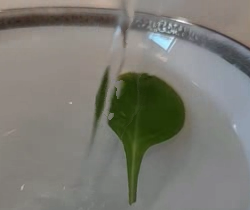}
  \subcaption{premiere-warp}
  \end{subfigure}
  \begin{subfigure}[b]{\fivewidth\linewidth}
  \includegraphics[width=\linewidth]{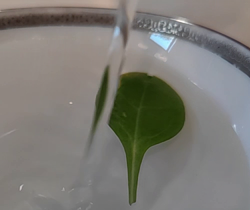}
  \subcaption{\textbf{Ours}}
  \end{subfigure}
\end{center} 
\vspace{-3mm}
\caption{Comparison of the intermediate frame synthesized by different methods. The blending result has obvious ghosting artifacts (transparent edges), while the optical-flow warping result has holes in the leaf. Compared with the other two methods, our result does not have these artifacts.} 
\label{fig:fps_conversion}
\vspace{-3mm}
\end{figure*}

To quantitatively evaluate the frame rate conversion on in-the-wild videos of our network, we took 120-fps video clips using a high-speed camera. In this way, we can get the input and corresponding ground truth for 20-fps and 24-fps frames by sampling every 6 or 5 frames, respectively. We measure the PSNR and SSIM of the output and show the results in Table~\ref{tab:fps_conversion}. We also take the outputs of several popular video editing tools and software for comparison: ffmpeg~\cite{tomar2006converting} (skip/duplicate frames), and Adobe Premiere Pro~\cite{adobepremierepro}. Adobe Premiere Pro has three options for frame rate conversion: frame sampling, frame blending, and optical flow warping. Among these compared methods, only ``blending" and ``warping" can synthesize new frames at the new timestamp. We show the synthesized frames in Fig.~\ref{fig:fps_conversion}.

From Table~\ref{tab:fps_conversion}, we can conclude that the frame duplication and sampling perform the worst since it cannot synthesize frames at the new timestamp. Frame blending is a little bit better in terms of PSNR and SSIM; still, it suffers from ghosting artifacts (transparent edges due to the overlayed objects) as shown in the second figure in Fig.~\ref{fig:fps_conversion}. Optical flow warping can synthesize the correct motion for the new timestamp and thus improve the reconstruction PSNR and SSIM. However, the reconstruction performance relies on the estimated optical flow field: if the estimation accuracy is low, then the warped frames would have holes, as shown in the third figure. Compared with the above methods, our network does not have these artifacts and achieves the highest reconstruction quality in terms of PSNR and SSIM.

\begin{table*}[tbp]
\caption{Performance comparison for frame rate conversion: 20 to 24 fps.}
\label{tab:fps_conversion}
\vspace{-3mm}
\begin{center}
\begin{tabular}{c|cccccc}
\hline
Method & ffmpeg~\cite{tomar2006converting} & premiere-sampling &	premiere-blend &	premiere-warp &	Ours \\ 
\hline 
PSNR & 32.44 &	32.39 &	33.90 &	34.54 &	\textbf{36.99}  \\
SSIM & 0.9514 &	0.9474 &	0.9536 &	0.9588 &	\textbf{0.9784}  \\
\hline
\end{tabular}
\end{center}
\vspace{-3mm}
\end{table*}


\subsection{Blurry Frame Reconstruction}
\noindent
\textbf{Training Details.} For this application, we adopt REDS~\cite{Nah_2019_CVPR_Workshops_REDS} as our training set: it provides blurry videos at 24-fps and the corresponding clean videos at both 24-fps or 120-fps. We conduct two experiments: in the main paper, we show the results of $4\times s, 2\times t$, which are trained with bicubic-downsampled blurry videos at 12-fps as input, and clean HR videos at 24-fps as output. Besides, we also experiment on another setting with temporal-upscaling only: $1\times s, 5\times t$, which means the model needs to decode 5 clean frames per input frame. This network is trained with 24-fps blurry videos as input and 120-fps clean videos as output. To handle the motion ambiguity, we take 2 frames as input and make the network generate $2\times t$ frames as output. We use the training subset of REDS to train the networks and show results on the validation subset.

\begin{figure*}[htbp]
\captionsetup[subfigure]{labelformat=empty}
\begin{center}
  \begin{subfigure}[b]{\fivewidth\linewidth}
  \includegraphics[width=\linewidth]{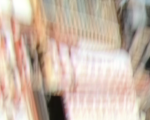}
  \subcaption{Input}
  \end{subfigure}
  \begin{subfigure}[b]{\fivewidth\linewidth}
  \includegraphics[width=\linewidth]{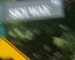}
    \subcaption{Input}
  \end{subfigure}

  \begin{subfigure}[b]{\fivewidth\linewidth}
  \includegraphics[width=\linewidth]{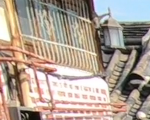}
  \end{subfigure}
  \begin{subfigure}[b]{\fivewidth\linewidth}
  \includegraphics[width=\linewidth]{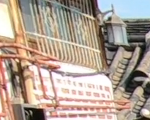}
  \end{subfigure}
  \begin{subfigure}[b]{\fivewidth\linewidth}
  \includegraphics[width=\linewidth]{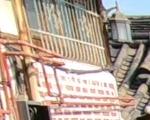}
  \end{subfigure}
  \begin{subfigure}[b]{\fivewidth\linewidth}
  \includegraphics[width=\linewidth]{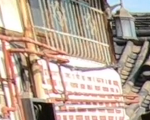}
  \end{subfigure}
  \begin{subfigure}[b]{\fivewidth\linewidth}
  \includegraphics[width=\linewidth]{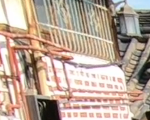}
  \end{subfigure}  
  
  \begin{subfigure}[b]{\fivewidth\linewidth}
  \includegraphics[width=\linewidth]{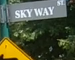}
  \subcaption{GT-0}
  \end{subfigure}
  \begin{subfigure}[b]{\fivewidth\linewidth}
  \includegraphics[width=\linewidth]{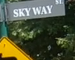}
  \subcaption{GT-1}
  \end{subfigure}
  \begin{subfigure}[b]{\fivewidth\linewidth}
  \includegraphics[width=\linewidth]{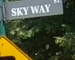}
  \subcaption{GT-2}
  \end{subfigure}
  \begin{subfigure}[b]{\fivewidth\linewidth}
  \includegraphics[width=\linewidth]{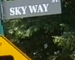}
  \subcaption{GT-3}
  \end{subfigure}
  \begin{subfigure}[b]{\fivewidth\linewidth}
  \includegraphics[width=\linewidth]{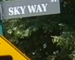}
  \subcaption{GT-4}
  \end{subfigure}

  \begin{subfigure}[b]{\fivewidth\linewidth}
  \includegraphics[width=\linewidth]{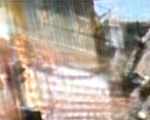}
  \end{subfigure}
  \begin{subfigure}[b]{\fivewidth\linewidth}
  \includegraphics[width=\linewidth]{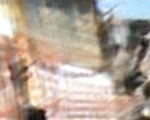}
  \end{subfigure}
  \begin{subfigure}[b]{\fivewidth\linewidth}
  \includegraphics[width=\linewidth]{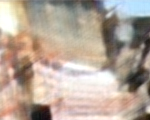}
  \end{subfigure}
  \begin{subfigure}[b]{\fivewidth\linewidth}
  \includegraphics[width=\linewidth]{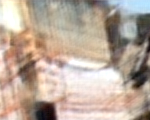}
  \end{subfigure}
  \begin{subfigure}[b]{\fivewidth\linewidth}
  \includegraphics[width=\linewidth]{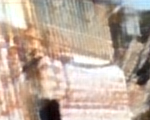}
  \end{subfigure}  
  
  \begin{subfigure}[b]{\fivewidth\linewidth}
  \includegraphics[width=\linewidth]{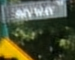}
  \subcaption{FLAVR-0}
  \end{subfigure}
  \begin{subfigure}[b]{\fivewidth\linewidth}
  \includegraphics[width=\linewidth]{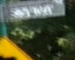}
  \subcaption{FLAVR-1}
  \end{subfigure}
  \begin{subfigure}[b]{\fivewidth\linewidth}
  \includegraphics[width=\linewidth]{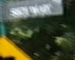}
  \subcaption{FLAVR-2}
  \end{subfigure}
  \begin{subfigure}[b]{\fivewidth\linewidth}
  \includegraphics[width=\linewidth]{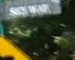}
  \subcaption{FLAVR-3}
  \end{subfigure}
  \begin{subfigure}[b]{\fivewidth\linewidth}
  \includegraphics[width=\linewidth]{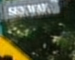}
  \subcaption{FLAVR-4}
  \end{subfigure}

  \begin{subfigure}[b]{\fivewidth\linewidth}
  \includegraphics[width=\linewidth]{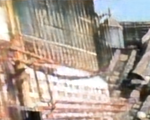}
  \end{subfigure}
  \begin{subfigure}[b]{\fivewidth\linewidth}
  \includegraphics[width=\linewidth]{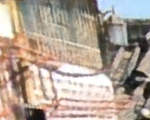}
  \end{subfigure}
  \begin{subfigure}[b]{\fivewidth\linewidth}
  \includegraphics[width=\linewidth]{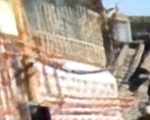}
  \end{subfigure}
  \begin{subfigure}[b]{\fivewidth\linewidth}
  \includegraphics[width=\linewidth]{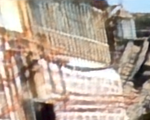}
  \end{subfigure}
  \begin{subfigure}[b]{\fivewidth\linewidth}
  \includegraphics[width=\linewidth]{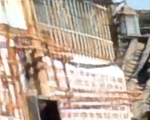}
  \end{subfigure}  
  
  \begin{subfigure}[b]{\fivewidth\linewidth}
  \includegraphics[width=\linewidth]{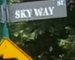}
  \subcaption{Ours-0}
  \end{subfigure}
  \begin{subfigure}[b]{\fivewidth\linewidth}
  \includegraphics[width=\linewidth]{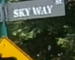}
  \subcaption{Ours-1}
  \end{subfigure}
  \begin{subfigure}[b]{\fivewidth\linewidth}
  \includegraphics[width=\linewidth]{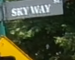}
  \subcaption{Ours-2}
  \end{subfigure}
  \begin{subfigure}[b]{\fivewidth\linewidth}
  \includegraphics[width=\linewidth]{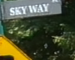}
  \subcaption{Ours-3}
  \end{subfigure}
  \begin{subfigure}[b]{\fivewidth\linewidth}
  \includegraphics[width=\linewidth]{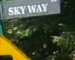}
  \subcaption{Ours-4}
  \end{subfigure}
\end{center} 
\caption{Qualitative results of the blurry frame reconstruction on REDS. The first row is the input blurry frame, and the other rows are corresponding clean frames. Our method can generate clean frames with vivid textures and correct motions at each timestep.} 
\label{fig:dfm_comp}
\end{figure*}

To further illustrate the effectiveness of our upsampler design, We compare our upsampler's structure with FLAVR~\cite{kalluri2020flavr} under the setting $1\times s, 5\times t$ in Table~\ref{tab:bin_dfm} trained with the same hyperparameters. Our method exceeds it by a large margin in terms of PSNR and SSIM. We also provide the qualitative results of the validation videos from REDS in Fig.~\ref{fig:dfm_comp}. As we can see, our method can reduce the motion blur and produce clean and crisp frames. Compared with FLAVR, our network generates vivid textures for each frame while keeping a good motion pattern at each timestep.

\begin{table}[tbp]
\caption{Comparison for blurry frame upscaling: $1\times s, 5\times t$.}
\label{tab:bin_dfm}
\begin{center}
\begin{tabular}{c|cc}
\hline
 \multirow{2}{*}{Method} & \multicolumn{2}{c}{REDS} \\ 
          & PSNR        &     SSIM                      \\ \hline
FLAVR~\cite{kalluri2020flavr} & 28.50 &	0.8337 \\
Ours & \textbf{32.29} &	\textbf{0.9153}
\\ \hline
\end{tabular}
\end{center}
\vspace{-4mm}
\end{table}

\begin{figure}[tbp]
\captionsetup[subfigure]{labelformat=empty}
\begin{center}
  \begin{subfigure}[b]{0.92\linewidth}
  \includegraphics[width=\linewidth]{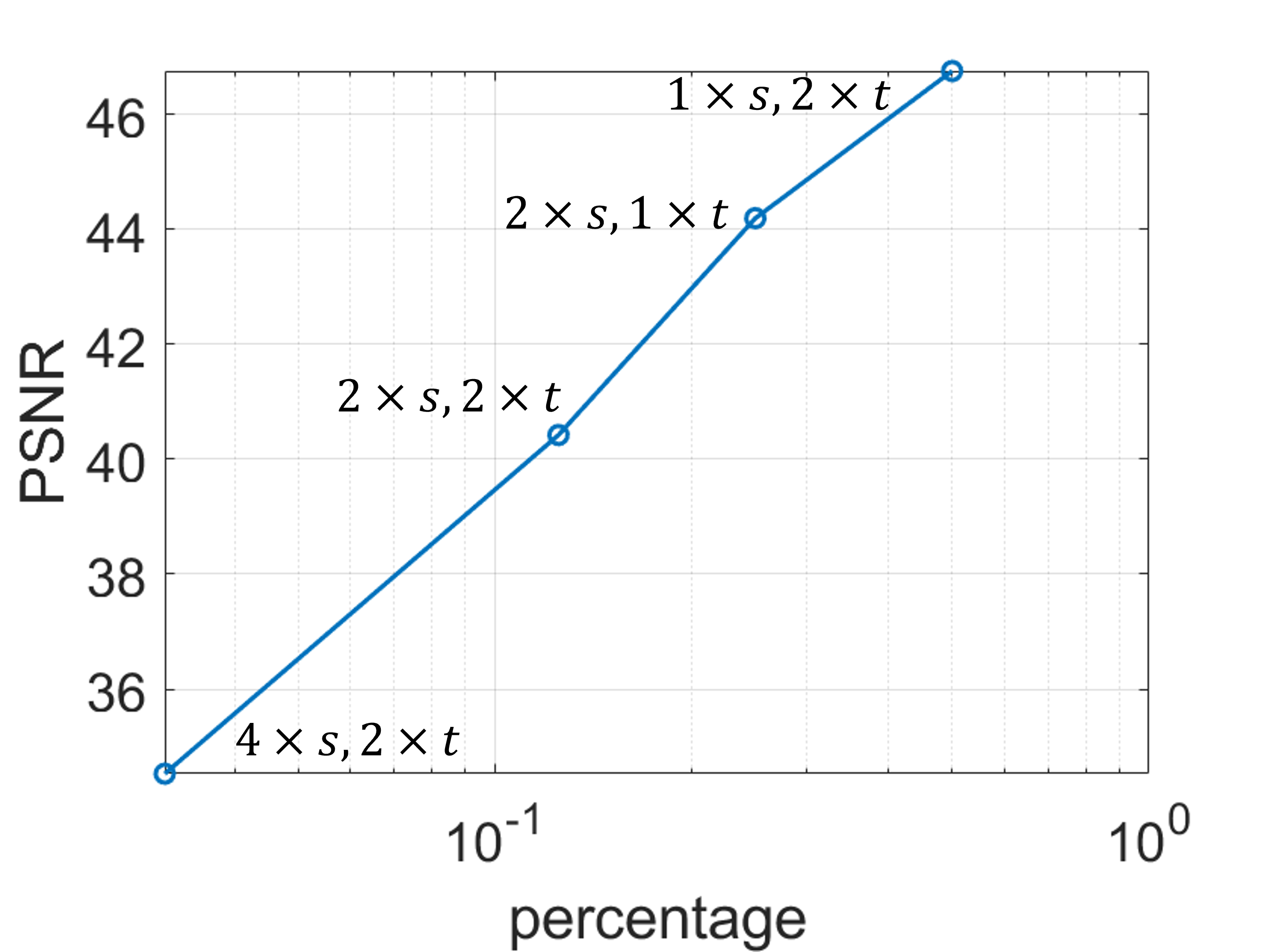}
  \subcaption{(a) PSNR}
  \end{subfigure}
  
 \begin{subfigure}[b]{0.92\linewidth}
 \includegraphics[width=\linewidth]{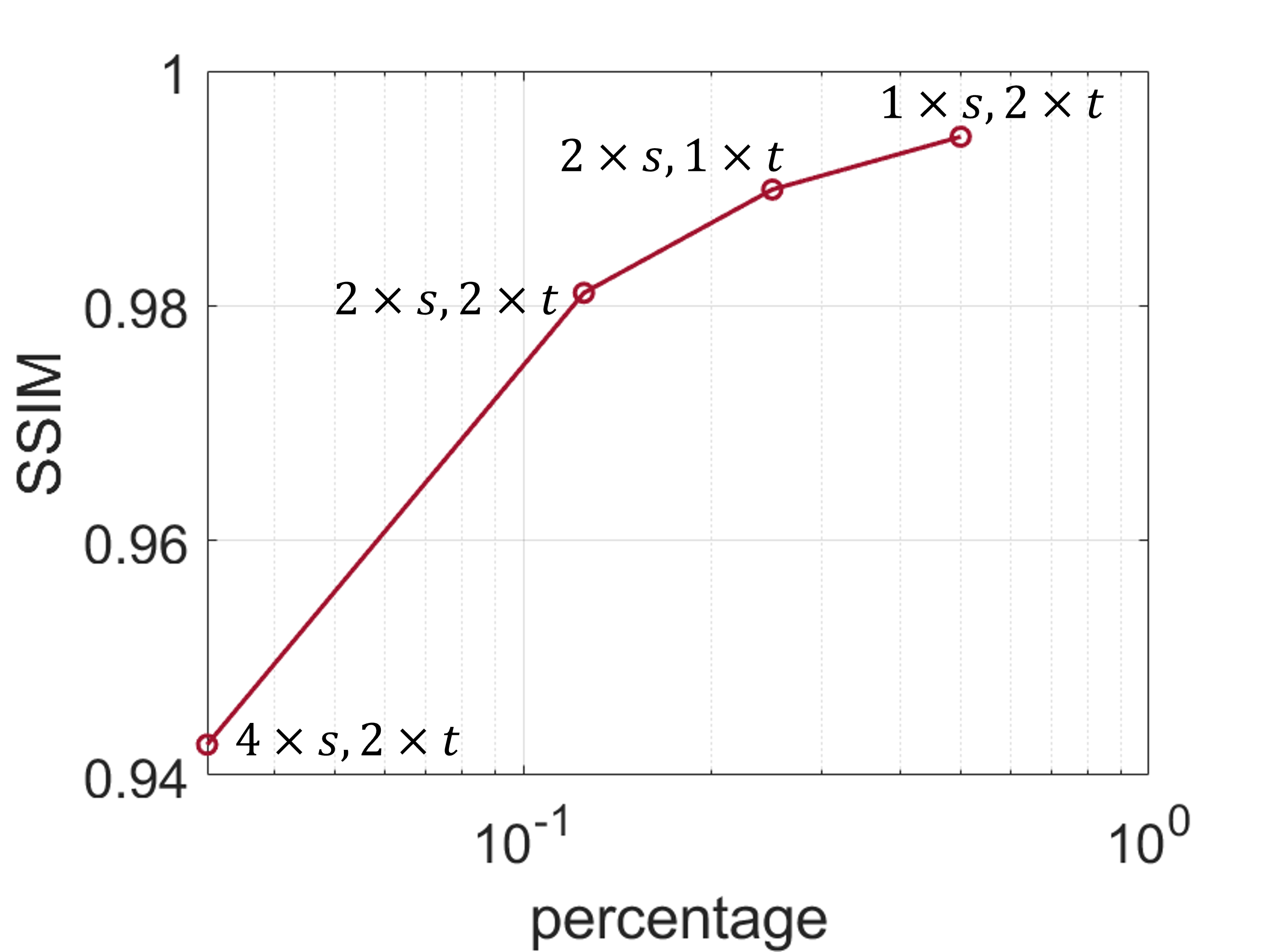}
  \subcaption{(b) SSIM}
  \end{subfigure}
\end{center}
\vspace{-3mm} 
\caption{\textit{Plot of reconstruction PSNR/SSIM for different percentages of pixels in the downsampled representation}. With the increase of pixel percentage, the reconstruction results improves.}
\vspace{-3mm}
\label{fig:trade_off}
\end{figure}

\subsection{Efficient Video Storage and Transmission}

For this application, we assume the video capturing and reconstruction steps are asynchronous: the captured video can be downsampled and saved or transmitted for later reconstruction. In this process, the downsampling filter can help save the data to be stored or transmitted while keeping high-quality reconstruction results. Our proposed downsampling method is a better way to reduce the number of pixels in space and time dimensions, and it is compatible with the previous image and video compression methods in the standard ISP pipeline. In practical applications, a video can be first resized by our STAA downsampler and then compressed with video codecs for transmission. Correspondingly, it requires the upsampler to be trained with the compressed data for better handling the compressed video reconstruction. We do not conduct the above experiments since it is beyond the scope of this paper. It can be a promising direction for future works.

For efficient video storage and transmission, we care about the downsampling ratio since it directly influences the amount of data to be stored or transmitted: intuitively, the restoration quality should increase with the sampling ratio. In the extreme case when the ratio = 1.0, the stored/transmitted video is exactly the original input. We discuss the trade-off between the data storage and reconstruction performance as follows.

\noindent
\textbf{Trade-off between Storage and Reconstruction Performance.} We evaluate the percentage of pixels kept in the downsampled representation: for temporal downsampling ratio of $t$, there will be $1/t$ pixels kept; for spatial downsampling ratio of $s$, there will be $1/s^2$ pixels kept because of the reduction in both height and width. We conduct a series of experiments with different combinations of space and time downsampling factors and plot the influence of \textit{pixel percentage in downsampled representation} to the \textit{reconstruction PSNR/SSIM} calculated on Vimeo-90k's testset in Fig.~\ref{fig:trade_off}, where the x-axis is the base-10 logarithmic scale of the pixel percentage, and the y-axis refers to PSNR and SSIM. All the experiments are done with the same network structure and trained with the same hyperparameters on Vimeo-90k.

From plot (a), we can observe that the reconstruction PSNR is almost linearly correlated to the log of pixel percentage. The SSIM plot increases slower when the percentage becomes higher, and it will eventually reach the top-right corner when the percentage is $100\%$ and SSIM=1.0 (original video). These plots reflect some interesting facts:

1. Under our STAA framework with jointly learned downsampler and upsampler, the reconstruction performance only relates to the percentage of pixels kept in the downsampled representation, regardless of whether the reduction happens on time or space dimension. This trend validates our method's effectiveness in maintaining the spatio-temporal characteristics of the video;

2. With this plot, we can estimate a rough range of reconstruction performance given a specific downsampling ratio. This trade-off can guide the design of the video restoration system by answering the following question: what is the lowest percentage of pixels that satisfies the reconstruction quality requirement;

3. This plot can serve as a benchmark of the network's restoration capability: a better network should move the plot to the top left (reconstruct the best quality with the least percentage of pixels). Since the plot is acquired through different space/time downsampling settings, it reflects the general restoration ability beyond specific tasks.

\end{appendices}
\end{document}